\documentclass{clv3}

\makeatletter
\jname{}
\renewcommand{\footmark}{}
\makeatother

\usepackage{etex}
\usepackage{fixltx2e}

\usepackage[utf8]{inputenc}
\usepackage[T1]{fontenc}
\usepackage[USenglish]{babel}

\usepackage{xcolor}

\usepackage{amsmath}
\usepackage{booktabs}
\usepackage{graphicx}
\usepackage[slantedGreek]{mathpazo}
\usepackage[defblank]{paralist}
\usepackage{proof}
\usepackage{tikz}
\usetikzlibrary{arrows}
\usetikzlibrary{calc}
\usetikzlibrary{matrix}
\usetikzlibrary{shapes.multipart}
\usetikzlibrary{positioning}
\usepackage{wasysym}
\usepackage{url}
\usepackage{ccg}

\makeatletter
\@ifundefined{useshorthands}{}{%
  \useshorthands{"}%
  \defineshorthand{"-}{\nobreak-\bbl@allowhyphens}%
  \defineshorthand{"=}{\nobreak--\bbl@allowhyphens}%
  \defineshorthand{""}{\nobreak\-\bbl@allowhyphens}%
}
\makeatother

\usepackage{hyperref}
\usepackage{xcolor}
\definecolor{darkblue}{rgb}{0, 0, 0.5}
\hypersetup{colorlinks=true,citecolor=darkblue, linkcolor=darkblue, urlcolor=darkblue}

\makeatletter
\newcommand{\A}{{\mathcal A}} 

\newcommand*{\SHORTCUTR}[1]{%
  \setbox\@tempboxa\vbox{\hbox{$\ccgstrut$}\kern\inferLineSkip\kern.4\p@\kern\inferLineSkip}%
  \@tempdima=\ht\@tempboxa
  \advance\@tempdima\dp\@tempboxa
  \multiply\@tempdima by 2
  \ifx\\#1\\\else
  \rlap{\quad\vbox to\@tempdima{\vfil\hbox{\small\strut #1}\vfil}}%
  \fi
  \vbox to\@tempdima{\leaders\hbox{\vphantom{x}.}\vfil\hbox{\vphantom{x}}}%
}
\newcommand*{\SHORTCUTL}[1]{%
  \setbox\@tempboxa\vbox{\hbox{$\ccgstrut$}\kern\inferLineSkip\kern.4\p@\kern\inferLineSkip}%
  \@tempdima=\ht\@tempboxa
  \advance\@tempdima\dp\@tempboxa
  \multiply\@tempdima by 2
  \ifx\\#1\\\else
  \llap{\vbox to\@tempdima{\vfil\hbox{\small\strut #1}\vfil}\quad}%
  \fi
  \vbox to\@tempdima{\leaders\hbox{\vphantom{x}.}\vfil\hbox{\vphantom{x}}}%
}
\newcommand*{\BC}[1][]{{<}\ifx\\#1\\\else_{#1}\fi}

\newcommand*{\CATEGORIES}[1]{\mathcal{C}(#1)}
\newcommand*{\CBOX}[2]{\hbox to#1{\hfill$\mathstrut #2$\hfill}}
\newcommand*{\CONTROL}[1]{[#1]}
\newcommand*{\EMPTYSTRING}{\varepsilon}

\newcommand*{\FC}[1][]{{>}\ifx\\#1\\\else_{#1}\fi}

\makeatletter
\newcommand*{\RDOTINFER}[3][]{\ensuremath{\@deduce{\@empty}[\SHORTCUTR{#1}]{\ccgstrut #2}{\ccgstrut #3}}}
\newcommand*{\LDOTINFER}[3][]{\ensuremath{\@deduce{\@empty}[\SHORTCUTL{#1}]{\ccgstrut #2}{\ccgstrut #3}}}
\makeatother
\newcommand*{\INST}{\Rrightarrow}
\newcommand*{\LEXICON}{\mathrel{:=}}
\newcommand*{\LIG}{LIG}

\newcommand*{\SET}[1]{\{#1\}}
\newcommand*{\SETC}[2]{\SET{#1\mid #2}}

\newcommand*{\STACK}{\mathinner{\$}}
\newcommand*{\TAG}{\VERSAL{TAG}}
\newcommand*{\TERM}[1]{\textbf{#1}}
\newcommand*{\TERMF}[1]{{#1}}

\newcommand*{\VWCCG}{\VERSAL{VW-CCG}}

\newcommand*{\HIGHLIGHT}[1]{%
  \bgroup\fboxsep=0pt\colorbox{yellow!60}{\mathstrut\ensuremath{#1}}\egroup%
}
\newcommand*{\TECH}{\cent}

\newcommand*{\ACCEPT}{\mathsf{A}}

\newcommand*{\ASSIGN}[2]{#1 \mapsto #2}
\newcommand*{\ASSIGNCAT}[2]{[\ASSIGN{#1}{#2}]}
\newcommand*{\CCG}{\VERSAL{CCG}}
\newcommand*{\CMD}[1]{\ensuremath{\text{\textbf{#1}}}}

\newcommand*{\EXPTIME}{\VERSAL{EXPTIME}}
\newcommand*{\FALSE}{\mathbf{0}}
\newcommand*{\FUN}[1]{\text{\textit{#1}}}
\newcommand*{\NEG}[1]{\overline{#1}}
\newcommand*{\NP}{\VERSAL{NP}}
\newcommand*{\PAD}[2][\quad]{#1#2#1}

\newcommand*{\REJECT}{\mathsf{R}}
\newcommand*{\RULE}[4][\RULESEP]{#2#1#3#1\Rightarrow#1#4}

\newcommand*{\SAT}{SAT}
\newcommand*{\TRUE}{\mathbf{1}}

\newcommand*{\VERSAL}[1]{#1}

\newcommand{\size}[1]{|#1|}
\newcommand{\eat}[1]{}

\newcommand{\myTM}{M}

\newcommand{\ep}{\varepsilon}

\newcommand{\order}[1]{{\cal O}({#1})}

\newcommand*{\ATOMIC}{{\cal A}}
\newcommand*{\LEXARGS}{{\cal L}}

\newcommand*{\PIMINUS}{\CONTROL{\pi; {-}}}
\newcommand*{\PIPLUS}{\CONTROL{\pi; {+}}}
\newcommand*{\PIEQ}[1]{\CONTROL{\pi; {=}_{#1}}}

\makeatother

\usepackage{microtype}

\begin{document}

\title{On the Complexity of CCG Parsing}

\author{Marco Kuhlmann\thanks{%
    Department of Computer and Information Science, Linköping
    University, 581\,83 Link\"oping,
    Sweden. E-mail:~\texttt{marco.kuhlmann@liu.se}}}
\affil{Linköping University}

\author{Giorgio Satta\thanks{%
    Department of Information Engineering, University of Padua, via
    Gradenigo 6/A, 35131 Padova,
    Italy. E-mail:~\texttt{satta@dei.unipd.it}}}
\affil{University of Padua}

\author{Peter Jonsson\thanks{%
    Department of Computer and Information Science, Linköping
    University, 581\,83 Link\"oping,
    Sweden. E-mail:~\texttt{peter.jonsson@liu.se}}}
\affil{Linköping University}

\historydates{%
  Submission received: February 21, 2017;
  revised submission received: ?;\\
  accepted for publication: ?.}

\runningauthor{Kuhlmann, Satta, Jonsson}
\runningtitle{On the Complexity of CCG Parsing}

\maketitle

\begin{abstract}
  We study the parsing complexity of Combinatory Categorial Grammar
  (\CCG) in the formalism of \citet{vijayshanker1994equivalence}.
  As our main result, we prove that any parsing algorithm for this
  formalism will take in the worst case exponential time when the size
  of the grammar, and not only the length of the input sentence, is
  included in the analysis.
  This sets the formalism of \citet{vijayshanker1994equivalence} apart
  from weakly equivalent formalisms such as Tree"-Adjoining Grammar
  (\TAG), for which parsing can be performed in time polynomial in the
  combined size of grammar and input sentence.
  Our results contribute to a refined understanding of the class of
  mildly context"-sensitive grammars, and inform the search for new,
  mildly context"-sensitive versions of \CCG.
\end{abstract}

\section{Introduction}
\label{sec:Introduction}

Combinatory Categorial Grammar (\CCG;
\citealp{steedman2011combinatory}) is a well"-established grammatical
framework that has supported a large amount of work both in linguistic
analysis and natural language processing.
From the perspective of linguistics, the two most prominent features
of \CCG\ are its tight coupling of syntactic and semantic information,
and its capability to compactly encode this information entirely
within the lexicon.
Despite the strong lexicalization that characterizes \CCG, it is able
to handle non"-local dependencies in a simple and effective way
\citep{rimell2009unbounded}.
After the release of annotated datasets
\citep{hockenmaier2007ccgbank}, there has been a surge of interest in
\CCG\ within statistical and, more recently, neural natural language
processing.
The wide range of applications for which \CCG\ has been used includes
data"-driven syntactic parsing \citep{clark2007wide,zhang2011shift},
natural language generation
\citep{white2010generating,zhang2015discriminative}, machine
translation \citep{Lewis-emnlp-2013}, and broad"-coverage semantic
parsing \citep{Lewis14a*ccg,Lee-emnlp-2016}.

In this article we study the parsing complexity of \CCG.
Our point of departure is the work of
\citet{vijayshanker1990polynomial}, who presented the first
polynomial"-time parsing algorithm for \CCG.
The runtime complexity of this algorithm is in $\order{n^6}$, where
$n$ is the length of the input sentence.
This matches the runtime complexity of standard parsing algorithms for
Tree"-Adjoining Grammar (\TAG, \citealp{schabes1990mathematical}),
which fits nicely with the celebrated result that \CCG\ and \TAG\ are
weakly equivalent
\citep{weir1988combinatory,vijayshanker1994equivalence}.
However, while the runtime of
\citeauthor{vijayshanker1990polynomial}'s algorithm is polynomial in
the length of the input sentence, it is \emph{exponential} in the size
of the grammar.
This is in contrast with the situation for \TAG, where the runtime is
(roughly) quadratic with respect to grammar size
\citep{schabes1990mathematical}.
The only other polynomial"-time parsing algorithms for \CCG\ that we
are aware of \citep{vijayshanker1993parsing,kuhlmann2014new} exhibit
the same behaviour.
\citet{kuhlmann2014new} ask whether parsing may be inherently more
complex for \CCG\ than for \TAG\ when grammar size is taken into
account.
Our main technical result in this article is that the answer to this
question is positive: We show that \emph{any} parsing algorithm for
\CCG\ in the formalism considered by
\citeauthor{vijayshanker1990polynomial} will \emph{necessarily} take
in the worst case exponential time when the size of the grammar is
included in the analysis.
Formally, we prove that the universal recognition problem for this
formalism is \EXPTIME"-complete.
The following paragraphs provide some context to this result.

\paragraph{The Mild Context"-Sensitivity of Modern \CCG}

Our interest in the computational properties of \CCG\ is motivated by
our desire to better understand modern incarnations of this framework
from a mathematical point of view.
Theoretical work on \CCG\ has always emphasized the importance of
keeping the computational and generative power of the grammar as low
as possible (see for instance \citealp[p.~23]{steedman2000syntactic},
and \citealp[Section~2.5]{baldridge2002lexically}), and in doing so
has followed the tradition of the so"-called mildly context"-sensitive
theories of grammar.
The aforementioned polynomial"-time parsing algorithm and the weak
equivalence with \TAG\ established the membership of \CCG\ in this
class of grammars even on a formal level.
However, recent work has drawn attention to the fact that the specific
formalism for which these results were obtained, and which we will
refer to as \VWCCG\ (after Vijay"-Shanker and Weir), differs from
contemporary versions of \CCG\ in several important aspects.
In particular, it allows to restrict and even ban the use of
combinatory rules on a per"-grammar basis, whereas modern \CCG\
postulates one universal set of rules, controlled by a fully
lexicalized mechanism based on typed slashes, as in other approaches
to categorial grammar
\citep{baldridge2002lexically,steedman2011combinatory}.
The difference is important because the weak equivalence result
crucially depends on the availability of grammar"-specific rule
restrictions---without this feature, the generative power of \VWCCG\
is strictly smaller than that of \TAG\
\citep{kuhlmann2015lexicalization}.
At the same time, modern \CCG\ includes combinatory rules that are
absent from \VWCCG, specifically substitution and type"-raising, and
there is the possibility that this can counterbalance the loss of
generative power that comes with the lexicalization of the rule
control mechanism.
Then again, these new rules are not supported by existing
polynomial"-time parsing algorithms.
Moreover, the weak equivalence proof uses another feature of \VWCCG\
that is not available in contemporary versions of \CCG: the ability to
assign lexicon entries to the empty string.
Such ``empty categories'' are ruled out by one of the fundamental
linguistic principles of \CCG, the Principle of Adjacency
\cite[p.~54]{steedman2000syntactic}, and it is far from obvious that
the weak equivalence proof can be re"-written without them.
In summary, the formalism of Vijay"-Shanker and Weir is the only \CCG\
formalism that has been proved to be weakly equivalent to
\TAG,\footnote{\citet{baldridge2003multimodal} show that the weak
  generative power of their formalism for multi"-modal \CCG\ is
  \emph{at most as} strong as that of \TAG, but they do not show that
  it is \emph{at least as} strong.} and the only one that has been
shown to be parsable in polynomial time.
As such, it is arguably the only \CCG\ formalism that has been shown
to be mildly context"-sensitive, which is why we consider it to be of
continued interest from a mathematical point of view.
At the same time, we hope that the insights that we can obtain from
the analysis of \VWCCG\ will eventually lead to the development of
linguistically more adequate, provably mildly context"-sensitive
formalisms for \CCG.

\paragraph{Universal Recognition}

The \emph{universal recognition problem} for a class of grammars
$\mathcal{G}$ is the problem defined as follows: Given as input a
grammar $G$ in~$\mathcal{G}$ and a string $w$, decide whether $w$ is
in $L(G)$, the language generated by~$G$.
The computational complexity of this problem is measured as a function
of the combined size of $G$ and~$w$.
The universal recognition problem should be contrasted with the
\emph{membership problem} for any specific grammar $G$
in~$\mathcal{G}$, whose complexity is measured as a function solely of
the length of~$w$.
The complexity of the universal recognition problem is generally
higher than that of the membership problem.
For instance, the universal recognition problem for context"-free
grammars is PTIME"-complete (complete for decision problems solvable
in deterministic polynomial time), whereas the membership problem for
these grammars defines the class LOGCFL (decision problems reducible
in logarithmic space to a context"-free language), which is generally
conjectured to be a proper subset of PTIME.

The definitions of the universal recognition problem and the
membership problem often generate some confusion. 
For instance, in applications such as parsing or translation, we work
with a \emph{fixed} grammar, so it might seem that the universal
recognition problem is of little practical relevance.
However, it is worth remembering that for these applications, we are
primarily interested in the structural descriptions that the grammar
assigns to a generated sentence, not in the membership of the sentence
\emph{per se}.
Therefore the universal recognition problem is a more accurate model
of parsing than the membership problem, as the latter also admits
decision procedures where the grammar is replaced with some other
mechanism that may produce no or completely different descriptions
than the ones we are interested in.
The universal recognition problem is also favoured when the ambition
is to characterize parsing time in terms of all relevant inputs---both
the length of the input string and the size and structure of the
grammar \citep{ristad1986computational}.
Such an analysis often reveals (and does so even in this article) how
specific features of the grammar contribute to the complexity of the
parsing task.
More precisely, when investigating the universal recognition problem
one expresses the computational complexity of parsing in terms of
several parameters (other than the input string length), as for
instance the number of nonterminals, maximum size of rules, or maximum
length of unary derivations.
This provides a much more fine"-grained picture than the one that we
get when analyzing the membership problem, and discloses the effects
that each individual feature of the grammar has on parsing.

\paragraph{Structure of the Article}

The remainder of this article is structured as follows.
After presenting the \VWCCG\ formalism in
Section~\ref{sec:Preliminaries}, we first study in
Section~\ref{sec:NPCompleteness} the universal recognition problem for
a restricted class of \VWCCG, where each category is ``lexicalized''
in the sense of the Principle of Adjacency.
We show that for this subclass, universal recognition is NP"-complete.
Under the assumption that PTIME $\neq$ NP, this already implies our
main result that parsing algorithms for \VWCCG\ will take in the worst
case exponential time in the combined size of the grammar and the
input string.
In Section~\ref{sec:EXPTIMECompleteness} we analyze the general case
and show that the universal recognition problem for unrestricted
\VWCCG\ is \EXPTIME"-complete.
This is a stronger result than the one in
Section~\ref{sec:NPCompleteness}, as it does not rely on any
assumptions.
However, we anticipate that many readers will be content with the
result in Section~\ref{sec:NPCompleteness}, especially since the
proofs of the more general result are considerably more complex.
Finally, Section~\ref{sec:Discussion} is devoted to a general
discussion of our results, its ramifications, and its relevance for
current research.

\section{Preliminaries}
\label{sec:Preliminaries}

In this section we present the \VWCCG\ formalism.
We assume the reader to be already familiar with the basic notions of
categorial grammar, and in particular with the idea of categories as
syntactic types.
Like other categorial formalisms, a \VWCCG\ grammar has two central
components: a \TERM{lexicon}, which specifies the categories for
individual words, and a set of \TERM{rules}, which specify how to
derive the categories of longer phrases from the categories of their
constituent parts.

\subsection{Lexicon}

The \VWCCG\ lexicon is a set of pairs $\sigma \LEXICON X$, where
$\sigma$ is a word (formalized as a symbol from some finite
vocabulary) and $X$ is a category.
Formally, the set of \TERM{categories} over a given set $\A$ is the
smallest set $\CATEGORIES{\A}$ such that
\begin{inparaenum}[(i)]
\item $\A \subseteq \CATEGORIES{\A}$ and
\item if $X \in \CATEGORIES{\A}$ and $Y \in \CATEGORIES{\A}$ then
  $X \SLASHF Y \in \CATEGORIES{\A}$ and
  $X \SLASHB Y \in \CATEGORIES{\A}$.
\end{inparaenum}
Categories of form (i) are called \TERM{atomic}, those of form (ii)
are called \TERM{complex}.

\paragraph{Categories as Stacks}

Categories are usually viewed as directed versions of the function
types in the simply"-typed lambda calculus.
Here we follow authors such as
\citet[p.~195f.]{baldridge2002lexically} and view them as
\emph{stacks}.
We treat slashes as left"-associative operators and omit unncessary
parentheses.
This lets us to write every category $X \in \CATEGORIES{\A}$ in the
form
\begin{displaymath}
  X \PAD[\;]{=} A \SLASH_1 X_1 \cdots \SLASH_m X_m
\end{displaymath}
where $m \geq 0$, $A \in \A$ is an atomic category that we call the
\TERM{target} of~$X$, and the $\SLASH_i X_i$ are slash"=category pairs
that we call the \TERM{arguments} of~$X$.
Based on this notation we view $X$ as a pair consisting of the target
$A$ and a stack whose elements are the arguments of~$X$, with the
argument $\SLASH_m X_m$ at the top of the stack.
Note that arguments can in general contain complex categories.

\begin{figure}
  \small\vskip-\abovedisplayskip
  \begin{displaymath}
    \INFER[\eqref{rule:BackwardApplication}]{\text{S}}{%
      \PROJECT[3]{\text{NP}}{\text{We}}
      &
      \INFER[\eqref{rule:ForwardApplication}]{\text{S} \SLASHB \text{NP}}{%
        \PROJECT[2]{(\text{S} \SLASHB \text{NP}) \SLASHF \text{NP}}{\text{prove}}
        &
        \INFER[\eqref{rule:ForwardApplication}]{\text{NP}}{%
          \PROJECT{\text{NP} \SLASHF \text{N}}{\text{two}}
          &
          \PROJECT{\text{N}}{\text{theorems}}
        }
      }
    }
  \end{displaymath}
  \caption{A sample derivation tree for the sentence ``We prove two
    theorems''.}
  \label{fig:ExampleDerivation}
\end{figure}

\begin{example}
  In the derivation shown in Figure~\ref{fig:ExampleDerivation},
  lexicon assignments are indicated by dotted lines: The verb
  \emph{prove} for example is associated with the complex category
  $(\text{S} \SLASHB \text{NP}) \SLASHF \text{NP}$.
  In our notation, the same category can also be written as
  $\text{S} \SLASHB \text{NP} \SLASHF \text{NP}$.
  The target of this category is $\text{S}$, and the stack consists of
  the two arguments $\SLASHB \text{NP}$ and $\SLASHF \text{NP}$, with
  $\SLASHF \text{NP}$ at the top of the stack.
  Note that we follow the standard convention in \CCG\ and draw
  derivations with the leaves at the top and the root at the bottom.
\end{example}

\subsection{Rules}
\label{sec:rules}

\begin{figure}
  \small\vskip-\abovedisplayskip
  \begin{align}
    \RULE{X \SLASHF Y}{Y &}{X} && \text{(forward application)} \label{rule:ForwardApplication}
    \\
    \RULE{Y}{X \SLASHB Y &}{X} && \text{(backward application)} \label{rule:BackwardApplication}
    \\[\medskipamount]
    \RULE{X \SLASHF Y}{Y \SLASHF Z &}{X \SLASHF Z} && \text{(forward harmonic composition)} \label{rule:ForwardHarmonicComposition}
    \\
    \RULE{X \SLASHF Y}{Y \SLASHB Z &}{X \SLASHB Z} && \text{(forward crossed composition)} \label{rule:ForwardCrossedComposition}
    \\
    \RULE{Y \SLASHB Z}{X \SLASHB Y &}{X \SLASHB Z} && \text{(backward harmonic composition)} \label{rule:BackwardHarmonicComposition}
    \\
    \RULE{Y \SLASHF Z}{X \SLASHB Y &}{X \SLASHF Z} && \text{(backward crossed composition)} \label{rule:BackwardCrossedComposition}
  \end{align}
  \caption{The combinatory schemata with degree $0$ (application; top)
    and $1$ (composition; bottom).}
  \label{fig:CombinatorySchemata}
\end{figure}

The rules of \VWCCG\ are derived from two \TERM{combinatory schemata},
\begin{align*}
  \RULE{\HIGHLIGHT{X \SLASHF Y}}{Y \SLASH_1 Z_1 \cdots \SLASH_d Z_d&}{X \SLASH_1 Z_1 \cdots \SLASH_d Z_d}
  && \text{(forward schema)}
  \\
  \RULE{Y \SLASH_1 Z_1 \cdots \SLASH_d Z_d}{\HIGHLIGHT{X \SLASHB Y}&}{X \SLASH_1 Z_1 \cdots \SLASH_d Z_d}
  && \text{(backward schema)}
\end{align*}
where $d \geq 0$, the $\SLASH_i$ are slashes (forward or backward),
and $X$, $Y$ and the $Z_i$ are variables ranging over categories.
Each schema specifies how to combine a \TERMF{primary input category}
(highlighted) and a \TERMF{secondary input category} into an
\TERMF{output category}.
The integer $d$ is called the \TERM{degree} of the schema.
Rules derived from the schemata where $d = 0$ are called
\TERM{application} rules; the others are called \TERM{composition}
rules.

\begin{example}
  Figure~\ref{fig:CombinatorySchemata} shows all (six) combinatory
  schemata with degree at most~1, together with their conventional
  names.
  In the derivation shown in Figure~\ref{fig:ExampleDerivation}, each
  branching of the tree is annotated with the schema used in that
  step.
\end{example}

A \TERM{combinatory rule} over a set of categories $\CATEGORIES{\A}$
is obtained from a combinatory schema by optionally restricting the
ranges of some of the variables.
Two types of restrictions are possible:
\begin{inparaenum}[(i)]
\item We may require the variable $Y$ or any of the $Z_i$ to take the
  value of some specific category in $\CATEGORIES{\A}$.
  For example, we could derive a restricted version of backward
  crossed composition that applies only if
  $Y = \text{S} \SLASHB \text{NP}$:
  \begin{equation}
    \RULE{(\text{S} \SLASHB \text{NP}) \SLASHF Z}{X \SLASHB (\text{S} \SLASHB \text{NP})}{X \SLASHF Z}
    \label{rule:RestrictedBackwardCrossedComposition1}
  \end{equation}
\item We may restrict the range of the variable $X$ to categories with
  a specific target $A \in \A$.
  For example, we could restrict backward crossed composition to apply
  only in situations where the target of $X$ is $\text{S}$, the
  category of complete sentences.
  We denote the resulting rule using the ``$\STACK$'' notation of
  \citet{steedman2000syntactic}, where the symbol $\STACK$ is used as
  a variable for the part of the category stack below the topmost
  stack element:
  \begin{equation}
    \RULE{Y \SLASHF Z}{\text{S} \STACK \SLASHB Y}{\text{S} \STACK \SLASHF Z}
    \label{rule:RestrictedBackwardCrossedComposition2}
  \end{equation}
\end{inparaenum}

\begin{figure}
  \footnotesize\vskip-\abovedisplayskip
  \begin{displaymath}
    \INFER[\eqref{rule:BackwardApplication}]{\text{S}}{%
      \PROJECT[5]{\text{NP}}{\text{\strut Kahn}}
      &
      \INFER[\eqref{rule:ForwardApplication}]{\text{S} \SLASHB \text{NP}}{%
        \INFER[\eqref{rule:BackwardCrossedComposition}]{(\text{S} \SLASHB \text{NP}) \SLASHF \text{NP}}{%
          \PROJECT[3]{(\text{S} \SLASHB \text{NP}) \SLASHF \text{NP}}{\text{\strut blocked}}
          &
          \PROJECT[3]{(\text{S} \SLASHB \text{NP}) \SLASHB (\text{S} \SLASHB \text{NP})}{\text{\strut skillfully}}
        }
        &
        \INFER[\eqref{rule:ForwardApplication}]{\text{NP}}{%
          \PROJECT[3]{\text{NP} \SLASHF \text{N}}{\text{\strut a}}
          &
          \INFER[\eqref{rule:BackwardApplication}]{\text{N}}{%
            \INFER[\eqref{rule:ForwardApplication}]{\text{N}}{%
              \PROJECT{\text{N} \SLASHF \text{N}}{\text{\strut powerful}}
              &
              \PROJECT{\text{N}}{\text{\strut shot}}
            }
            &
            \INFER[\eqref{rule:ForwardApplication}]{\text{N} \SLASHB \text{N}}{%
              \PROJECT{(\text{N} \SLASHB \text{N}) \SLASHF \text{NP}}{\text{\strut by}}
              &
              \PROJECT{\text{NP}}{\text{\strut Rivaldo}}
            }
          }
        }
      }
    }
  \end{displaymath}
  \begin{displaymath}
    \INFER[\eqref{rule:BackwardApplication}]{\text{S}}{%
      \PROJECT[6]{\text{NP}}{\llap{*~}\text{Kahn}}
      &
      \INFER[\eqref{rule:ForwardApplication}]{\text{S} \SLASHB \text{NP}}{%
        \INFER[\eqref{rule:BackwardCrossedComposition}]{(\text{S} \SLASHB \text{NP}) \SLASHF \text{NP}}{%
          \PROJECT[4]{(\text{S} \SLASHB \text{NP}) \SLASHF \text{NP}}{\text{blocked}}
          &
          \PROJECT[4]{(\text{S} \SLASHB \text{NP}) \SLASHB (\text{S} \SLASHB \text{NP})}{\text{skillfully}}
        }
        &
        \INFER[\eqref{rule:ForwardApplication}]{\text{NP}}{%
          \PROJECT[4]{\text{NP} \SLASHF \text{N}}{\text{\strut a}}
          &
          \INFER[\eqref{rule:ForwardApplication}]{\text{N}}{%
            \INFER[\eqref{rule:BackwardCrossedComposition}\rlap{\ $\dagger$}]{\text{N} \SLASHF \text{N}}{%
              \PROJECT[2]{\text{N} \SLASHF \text{N}}{\text{\strut powerful}}
              &
              \INFER[\eqref{rule:ForwardApplication}]{\text{N} \SLASHB \text{N}}{%
                \PROJECT{(\text{N} \SLASHB \text{N}) \SLASHF \text{NP}}{\text{\strut by}}
                &
                \PROJECT{\text{NP}}{\text{\strut Rivaldo}}
              }
            }
            &
            \PROJECT[3]{\text{N}}{\text{\strut shot}}
          }
        }
      }
    }
  \end{displaymath}
  \caption{Overgeneration caused by unrestricted backward crossed composition.}
  \label{fig:OvergenerationBackwardCrossedComposition}
\end{figure}

\begin{example}
  Backward crossed composition \eqref{rule:BackwardCrossedComposition}
  can be used for the analysis of heavy NP shift in sentences such as
  \emph{Kahn blocked skillfully a powerful shot by Rivaldo} (example
  from \citealp{baldridge2002lexically}).
  A derivation for this sentence is shown in the upper half of
  Figure~\ref{fig:OvergenerationBackwardCrossedComposition}.
  However, the schema cannot be universally active in English, as this
  would cause the grammar to also accept strings such as \emph{*Kahn
    blocked skillfully a powerful by Rivaldo shot}, which is witnessed
  by the derivation in the lower half of
  Figure~\ref{fig:OvergenerationBackwardCrossedComposition} (a
  dagger~$\dagger$ marks the problematic step).
  To rule out this derivation, instead of the unrestricted schema, a
  \VWCCG\ grammar of English may select only certain instances of this
  schema as rules, in particular instances that combine the two
  restrictions in \eqref{rule:RestrictedBackwardCrossedComposition1}
  and \eqref{rule:RestrictedBackwardCrossedComposition2}.
  In this way the unwanted derivation in
  Figure~\ref{fig:OvergenerationBackwardCrossedComposition} can be
  blocked, while the other derivation is still admissible.
  Other syntactic phenomena require other grammar"-specific
  restrictions, including the complete ban of certain combinatory
  schemata (cf.\ \citealp[Sections
  4.2.1--4.2.2]{steedman2000syntactic}).
\end{example}

A \TERM{ground instance} of a combinatory rule over $\CATEGORIES{\A}$
is obtained by replacing every variable with a concrete category from
$\CATEGORIES{\A}$.
We denote ground instances using a triple arrow.
For example, the two instances of backward crossed composition in
Figure~\ref{fig:OvergenerationBackwardCrossedComposition} are:
\begin{displaymath}
  (\text{S} \SLASHB \text{NP}) \SLASHF \text{NP} \;\; 
  (\text{S} \SLASHB \text{NP}) \SLASHB (\text{S} \SLASHB \text{NP})
  \PAD[\;\;]{\INST}
  (\text{S} \SLASHB \text{NP}) \SLASHF \text{NP}
  \qquad\qquad
  \text{N} \SLASHF \text{N} \;\;
  \text{N} \SLASHB \text{N}
  \PAD[\;\;]{\INST} \text{N} \SLASHF \text{N}
\end{displaymath}
Every combinatory rule over $\CATEGORIES{\A}$ has infinitely many
ground instances.
In particular, the variable $X$ in such a rule can be replaced with
infinitely many concrete categories.

\subsection{Grammars}
\label{ssec:grammar}

A \VWCCG\ grammar fixes a finite lexicon and a finite set of
combinatory rules.
Formally, a grammar is defined as a structure
$G = (\Sigma, \A, {\LEXICON}, R, S)$ where $\Sigma$ is a finite
vocabulary, $\A$ is a finite set of atomic categories, ${\LEXICON}$ is
a finite relation between the sets
$\Sigma_\EMPTYSTRING = \Sigma \cup \SET{\EMPTYSTRING}$ and
$\CATEGORIES{\A}$, $R$~is a finite set of combinatory rules over
$\CATEGORIES{\A}$, and $S \in \A$ is a distinguished atomic category.
In what follows, we simply refer to the elements of~$R$ as rules.

\paragraph{Derivations}

Derivations of~$G$ are represented as binary trees whose nodes are
labeled with either lexicon entries (leaves) or categories (inner
nodes).
In order to represent such trees by linear terms, we use the following
notation.
Let $A$ be some unspecified alphabet.
For $a \in A$, the term $a$ represents a tree with a single node
labeled by~$a$.
For tree terms $t_1, \dots, t_m$, $m \geq 1$, the term
$a(t_1, \ldots, t_m)$ represents the tree whose root node is labeled
by $a$ and has $m$ children, which are the root nodes of the trees
represented by $t_1, \dots, t_m$.
With this notation, we define the set of \TERM{derivation trees}
of~$G$ and the associated mappings $\FUN{top}$ (which returns the
category at the root node of the tree) and $\FUN{yield}$ (which
returns the left"-to"-right concatenation of the symbols at the
leaves) recursively as follows:
\begin{itemize}
\item Every lexicon entry $\sigma \LEXICON X$ forms a (single"-node)
  derivation tree $\tau$.
  We define $\FUN{top}(\tau) = X$ and $\FUN{yield}(\tau) = \sigma$.
  \medskip
\item Let $\tau_L$ and $\tau_R$ be derivation trees with
  $\FUN{top}(\tau_L) = X_L$, $\FUN{yield}(\tau_L) = w_L$,
  $\FUN{top}(\tau_R) = X_R$, and $\FUN{yield}(\tau_R) = w_R$, and let
  $X_L \; X_R \INST X$ be a ground instance of some combinatory rule
  in~$R$.
  Then $\tau = X(X_L, X_R)$ is a derivation tree.
  We define $\FUN{top}(\tau) = X$ and $\FUN{yield}(\tau) = w_L w_R$,
  where juxtaposition denotes string concatenation.
\end{itemize}
The connection between this formal definition and the graphical
notation for derivation trees that we have used in
Figure~\ref{fig:ExampleDerivation} and
Figure~\ref{fig:OvergenerationBackwardCrossedComposition} should be
clear.
The only thing to note is that in a formal derivation tree, leaf nodes
correspond to lexicon entry $\sigma \LEXICON X$, while in our
graphical notation, leaf nodes are split into a parent node with the
category $X$ and a child, leaf node with the symbol~$\sigma$.

\paragraph{Generated Language}

Based on the concept of derivation trees, we can now define the string
language generated by a grammar~$G$.
The grammar~$G$ \TERMF{generates} a string $w$ if there exists a
derivation tree whose root node is labeled with the distinguished
category $S$ and whose yield equals~$w$.
The \TERM{language} generated by~$G$, denoted by $L(G)$, is the set of
all strings generated by~$G$.
As mentioned before, \citet{weir1988combinatory} and
\citet{vijayshanker1994equivalence} show \VWCCG\ generates the same
class of languages as Tree"-Adjoining Grammar (\TAG;
\citealp{joshi1997tree}).

\begin{figure}
  \footnotesize\vskip-\abovedisplayskip
  \begin{displaymath}
    \INFER[\eqref{rule:VijayShankerWeirExample2}]{S}{%
      \PROJECT[6]{A}{\mathstrut a}
      &
      \INFER[\eqref{rule:VijayShankerWeirExample2}]{S \SLASHB A}{%
        \PROJECT[5]{A}{\mathstrut a}
        &
        \INFER[\eqref{rule:VijayShankerWeirExample1}]{S \SLASHB A \SLASHB A}{%
          \INFER[\eqref{rule:VijayShankerWeirExample1}]{S \SLASHB A \SLASHB A \SLASHF \hat{S}}{%
            \INFER[\eqref{rule:VijayShankerWeirExample3}]{S \SLASHB A \SLASHB A \SLASHF \hat{S} \SLASHF B}{%
              \INFER[\eqref{rule:VijayShankerWeirExample1}]{S \SLASHB A \SLASHF \hat{S}}{%
                \PROJECT{S \SLASHB A \SLASHF \hat{S} \SLASHF B}{\mathstrut\EMPTYSTRING}
                &
                \PROJECT{B}{\mathstrut b}
              }
              &
              \PROJECT[2]{\hat{S} \SLASHB A \SLASHF \hat{S} \SLASHF B}{\mathstrut\EMPTYSTRING}
            }
            &
            \PROJECT[3]{B}{\mathstrut b}
          }
          &
          \PROJECT[4]{\hat{S}}{\EMPTYSTRING}
        }
      }
    }
  \end{displaymath}
  \begin{displaymath}
    \INFER[\eqref{rule:VijayShankerWeirExample2}]{S}{%
      \PROJECT[5]{A}{\mathstrut a}
      &
      \INFER[\eqref{rule:VijayShankerWeirExample1}]{S \SLASHB A}{%
        \INFER[\eqref{rule:VijayShankerWeirExample1}]{S \SLASHB A \SLASHF \hat{S}}{%
          \PROJECT[3]{S \SLASHB A \SLASHF \hat{S} \SLASHF B}{\mathstrut \EMPTYSTRING}
          &
          \PROJECT[3]{B}{\mathstrut b}
        }
        &
        \INFER[\eqref{rule:VijayShankerWeirExample2}]{\hat{S}}{%
          \PROJECT[3]{A}{\mathstrut a}
          &
          \INFER[\eqref{rule:VijayShankerWeirExample1}]{\hat{S} \SLASHB A}{%
            \INFER[\eqref{rule:VijayShankerWeirExample1}]{\hat{S} \SLASHB A \SLASHF \hat{S}}{%
              \PROJECT{\hat{S} \SLASHB A \SLASHF \hat{S} \SLASHF B}{\mathstrut\EMPTYSTRING}
              &
              \PROJECT{B}{\mathstrut b}
            }
            &
            \PROJECT[2]{\hat{S}}{\mathstrut\EMPTYSTRING}
          }
        }
      }
    }
  \end{displaymath}
  \caption{Two derivations of the grammar from Example~3.3 of
    \protect\citet{vijayshanker1994equivalence}.}
  \label{fig:VijayShankerWeirExample}
\end{figure}

\begin{example}\label{ex:VWG}
  \citet{vijayshanker1994equivalence} construct the following
  \VWCCG~$G$ (Example~3.3).
  We only specify the lexicon and the set of rules; the vocabulary,
  set of atomic categories and distinguished atomic category are left
  implicit.
  The lexicon is defined as follows:
  \begin{displaymath}
    a \LEXICON A\,,
    \quad
    b \LEXICON B\,,
    \quad
    \EMPTYSTRING \LEXICON S \SLASHB A \SLASHF \hat{S} \SLASHF B\,,
    \quad
    \EMPTYSTRING \LEXICON \hat{S} \SLASHB A \SLASHF \hat{S} \SLASHF B\,,
    \quad
    \EMPTYSTRING \LEXICON S\,,
    \quad
    \EMPTYSTRING \LEXICON \hat{S}
  \end{displaymath}
  The set of rules consists of all instances of application and all
  instances of forward composition of degree at most~$3$ where the
  target of the secondary input category is restricted to one of the
  ``hatted'' categories.
  We write $Y$ for a variable restricted to the set $\SET{S, A, B}$,
  $\hat{Y}$ for a variable restricted to the set
  $\SET{\hat{S}, \hat{A}, \hat{B}}$, and $Z_i$ for an unrestricted
  variable.
  As before, the $\SLASH_i$ are slashes (forward or backward).
  \stepcounter{equation}
  \begin{align}
    \RULE{X \SLASHF Y}{Y &}{X} \label{rule:VijayShankerWeirExample1}\\
    \RULE{Y}{X \SLASHB Y &}{X} \label{rule:VijayShankerWeirExample2}\\
    \RULE{X \SLASHF \hat{Y}}{\hat{Y} \SLASH_1 Z_1 \cdots \SLASH_d Z_d &}{X \SLASH_1 Z_1 \cdots \SLASH_d Z_d} && \text{where $0 \leq d \leq 3$} \label{rule:VijayShankerWeirExample3}
  \end{align}
  As witnessed by the derivations in
  Figure~\ref{fig:VijayShankerWeirExample}, the language generated by
  this grammar contains the subsets $\SETC{a^n b^n}{n \geq 0}$ and
  $\SETC{(ab)^n}{n \geq
    0}$.\footnote{\citet{vijayshanker1994equivalence} are mistaken in
    claiming that this grammar generates \emph{exactly}
    $\SETC{a^n b^n}{n \geq 0}$.}
\end{example}

The ability to impose restrictions on the applicability of rules plays
an important role in terms of generative power: without them, \VWCCG\
is strictly less powerful than \TAG\
\citep{kuhlmann2015lexicalization}.

\section{Complexity Without Categories for the Empty String}
\label{sec:NPCompleteness}

As already mentioned, the ability of \VWCCG\ to assign lexicon entries
to the empty string contradicts one of the central linguistic
principles of \CCG, the Principle of Adjacency, by which combinatory
rules may only apply to entities that are phonologically realized
\cite[p.~54]{steedman2000syntactic}.
In this section we therefore first investigate the computational
complexity of the universal recognition problem for a restricted
version of \VWCCG\ where this feature is dropped and every lexical
category is projected by an overt word.
We will say that a grammar $G$ whose lexicon does not contain any
assignments of the form $\EMPTYSTRING \LEXICON X$ is
\TERM{$\mathbold{\varepsilon}$"-free}.
We show the following result:

\begin{theorem}\label{thm:NPCompleteness}
  The universal recognition problem for $\EMPTYSTRING$"-free \VWCCG\
  is \NP"-complete.
\end{theorem}

We split the proof of this theorem into two parts:
Section~\ref{sec:NPHardness} shows hardness, and
Section~\ref{sec:NPMembership} shows membership.
Section~\ref{sec:NPDiscussion} contains a brief discussion of the
result.
For a gentle introduction to computational complexity and the relevant
proof techniques, we refer the reader to
\citet{papadimitriou1994computational}.

\subsection{NP"-Hardness}
\label{sec:NPHardness}

Our hardness proof is by a polynomial"-time reduction from the Boolean
Satisfiability Problem (\SAT) to the universal recognition problem for
$\EMPTYSTRING$"-free \VWCCG.
Since \SAT\ is an NP"-hard problem, this proves that the recognition
problem is NP"-hard as well.
An instance of \SAT\ is given by a Boolean formula $\phi$ in
conjunctive normal form.
This means that $\phi$ is a conjunction of clauses $c_i$, where each
clause consists of a disjunction of one or more literals.
A literal is either a variable $v_j$ or a negated variable
$\NEG{v_j}$.
The question asked about $\phi$ is whether it is satisfiable, that is,
whether there is a truth assignment to the variables that makes $\phi$
evaluate to $\TRUE$ (true).
Our reduction is a polynomial"-time procedure for transforming an
arbitrary instance $\phi$ into an $\EMPTYSTRING$"-free grammar~$G$ and
an input string~$w$ such that $\phi$ is satisfiable if and only if
$w \in L(G)$.
We additionally note that the combined size of~$G$ and $w$ is
polynomial in the total number of literals in~$\phi$, and thus obtain
the following:

\begin{lemma}\label{lem:NPHardness}
  The universal recognition problem for $\EMPTYSTRING$"-free \VWCCG\
  is NP"-hard.
\end{lemma}

We start with a description of how to obtain the input string~$w$ in
Section~\ref{sec:NPInputString}, and then turn to the grammar~$G$.
The lexicon and the rules of the grammar~$G$ will be set up in such a
way that every derivation for $w$ consists of three clearly separated
parts.
We will present these parts in sequence in sections
\ref{sec:NPHardnessPart1}--\ref{sec:NPHardnessPart3}, introducing the
relevant lexicon entries and rules as we go along.
The vocabulary and the set of atomic categories will be implicit.
We will say that we \emph{construct} the string $w$ and the grammar
$G$ based on~$\phi$.
Throughout the description of this construction we write $m$ for the
number of clauses in $\phi$ and $n$ for the total number of distinct
variables in~$\phi$.
The index $i$ will always range over values from~$1$ to~$m$ (clauses),
and the index $j$ will range over values from~$1$ to~$n$ (variables).
To illustrate our construction we will use the following instance of
SAT:
\begin{displaymath}
  \phi = (v_1 \lor \overline{v_2}) \land (v_1 \lor v_2) \land (\overline{v_1} \lor \overline{v_2})
\end{displaymath}
For this instance we have $m = 3$ and $n = 2$.
We can verify that the only truth assignment satisfying $\phi$ is
$\SET{\ASSIGN{v_1}{\TRUE}, \ASSIGN{v_2}{\FALSE}}$.
We set $c_1 = (v_1 \lor \overline{v_2})$, $c_2 = (v_1 \lor v_2)$, and
$c_3 = (\overline{v_1} \lor \overline{v_2})$.

\subsubsection{Input String}\label{sec:NPInputString}

We construct the input string as
\begin{displaymath}
  w = c_m \cdots c_1 c_0 v_1 \cdots v_n v_{n+1} d_n \cdots d_1
\end{displaymath}
where the $c_i$ and $v_j$ are symbols representing the clauses and
variables of the input formula~$\phi$, respectively.
The symbols $c_0$ and $v_{n+1}$ as well as the $d_j$ are special
symbols that we use for technical reasons, as explained below.
For our running example we have
$w = c_3 c_2 c_1 c_0 v_1 v_2 v_3 d_2 d_1$.

\subsubsection{Guessing a Truth Assignment}
\label{sec:NPHardnessPart1}

The first part of a derivation for~$w$ ``guesses'' a truth assignment
for the variables in~$\phi$ by assigning a complex category to the
substring $c_0 v_1 \cdots v_n v_{n+1}$.
Figure~\ref{fig:Guessing} shows how this could look like for our
running example.
Reading from the leaves to the root, for every symbol $v_j$ in $w$,
the derivation nondeterministically chooses between two lexicon
entries,
$v_j \LEXICON [\TECH] \SLASHF \ASSIGNCAT{v_j}{\TRUE} \SLASHF [\TECH]$ and
$v_j \LEXICON [\TECH] \SLASHF \ASSIGNCAT{v_j}{\FALSE} \SLASHF [\TECH]$;
these entries represent the two possible truth assignments to the
variable.
Note that we use square brackets to denote atomic categories.
The derivation then uses compositions \eqref{rule:Guessing1} and
\eqref{rule:Guessing2} to ``push'' these variable"-specific categories
to the argument stack of the lexical category for the special
symbol~$c_0$, and a final application \eqref{rule:Guessing3} to yield
a complex category that encodes the complete assignment.

\begin{figure}
  \vskip-\abovedisplayskip
  \begin{displaymath}
    \INFER[\eqref{rule:Guessing3}]{[c_0] \SLASHF \ASSIGNCAT{v_1}{\TRUE} \SLASHF 
      \ASSIGNCAT{v_2}{\FALSE}}{%
      \INFER[\eqref{rule:Guessing2}]{[c_0] \SLASHF \ASSIGNCAT{v_1}{\TRUE} \SLASHF 
        \ASSIGNCAT{v_2}{\FALSE} \SLASHF [\TECH]}{%
        \INFER[\eqref{rule:Guessing1}]{[c_0] \SLASHF \ASSIGNCAT{v_1}{\TRUE} \SLASHF [\TECH]}{%
          \PROJECT[1]{[c_0] \SLASHF [\TECH]}{c_0}
          &
          \PROJECT[1]{[\TECH] \SLASHF \ASSIGNCAT{v_1}{\TRUE} \SLASHF [\TECH]}{v_1}
        }
        &
        \PROJECT[2]{[\TECH] \SLASHF \ASSIGNCAT{v_2}{\FALSE} \SLASHF [\TECH] }{v_2}
      }
      &
      \PROJECT[3]{[\TECH]}{v_3}
    }
  \end{displaymath}
  \caption{Derivation fragment that ``guesses'' the truth assignment
    for the running example.}
  \label{fig:Guessing}
\end{figure}

\paragraph{Lexicon Entries and Rules}

To support derivations such as the one in Figure~\ref{fig:Guessing},
we introduce the following lexicon entries:
\begin{displaymath}
  c_0 \LEXICON [c_0] \SLASHF [\TECH]
  \qquad
  v_j \LEXICON [\TECH] \SLASHF \ASSIGNCAT{v_j}{\TRUE} \SLASHF [\TECH]
  \qquad
  v_j \LEXICON [\TECH] \SLASHF \ASSIGNCAT{v_j}{\FALSE} \SLASHF [\TECH]
  \qquad
  v_{n+1} \LEXICON [\TECH]
\end{displaymath}
We also introduce rules according to the following restricted
schemata.
Schemata~\eqref{rule:Guessing1} and~\eqref{rule:Guessing2} yield
composition rules of degree~$2$; schema~\eqref{rule:Guessing3} yields
application rules.
\begin{align}
  \RULE{X \SLASHF [\TECH]}{[\TECH] \SLASHF \ASSIGNCAT{v_j}{\TRUE} \SLASHF [\TECH]&}{X \SLASHF \ASSIGNCAT{v_j}{\TRUE} \SLASHF [\TECH]} \label{rule:Guessing1} 
  \\
  \RULE{X \SLASHF [\TECH]}{[\TECH] \SLASHF \ASSIGNCAT{v_j}{\FALSE} \SLASHF [\TECH]&}{X \SLASHF \ASSIGNCAT{v_j}{\FALSE} \SLASHF [\TECH]} \label{rule:Guessing2} 
  \\
  \RULE{X \SLASHF [\TECH]}{[\TECH]&}{X} \label{rule:Guessing3}
\end{align}

\paragraph{Derivational Ambiguity}

It is worth mentioning here that our rules support other derivation
orders than the left"-branching order shown in
Figure~\ref{fig:Guessing}.
In particular, we could first combine the variable"-specific
categories with each other, and then combine the result with the
category for~$c_0$.
One could rule out this derivational ambiguity by restricting the
target of the primary input of each of the rules above to the category
$[c_0]$, obtaining rules such as the following:\footnote{Recall from
  Section~\ref{sec:rules} that $\STACK$ is a variable for the part of
  the category stack below the topmost element.}
\begin{align}
  \RULE{[c_0] \STACK \SLASHF [\TECH]}{[\TECH] \SLASHF \ASSIGNCAT{v_j}{\TRUE} \SLASHF [\TECH]&}{[c_0] \STACK \SLASHF \ASSIGNCAT{v_j}{\TRUE} \SLASHF [\TECH]} \tag{\ref{rule:Guessing1}'}
\end{align}
For the purposes of our reduction, the different derivation orders are
irrelevant, and we therefore abstain from using target restrictions.

\subsubsection{Verifying the Truth Assignment}
\label{sec:NPHardnessPart2}

The second part of a derivation for $w$ verifies that the truth
assignment hypothesized in the first part satisfies all clauses.
It does so by using compositions to ``pass'' the stack of atomic
categories encoding the truth assignment from one clause to the next,
right"-to"-left.
For the running example, this could be done as in
Figure~\ref{fig:Verifying}.
Crucially, the rules used in this part are restricted in such a way
that the assignment can be ``passed'' to the next clause~$c_i$ only if
$c_i$ is satisfied by at least one assignment $\ASSIGN{v_j}{b}$.
This can happen in two ways: either the assignment sets $b = \TRUE$
and $v_j$ occurs in~$c_i$, or the assignment sets $b = \FALSE$ and the
negated variable $\overline{v_j}$ occurs in~$c_i$.
For example, the lowermost composition \eqref{rule:Select1} is
licensed because $v_1$ occurs in~$c_1$.
At the end of this part of the derivation, we have a complex category
encoding a truth assignment as before, but where we now also have
checked that this assignment satisfies all clauses.

\begin{figure}
  \vskip-\abovedisplayskip
  \begin{displaymath}
    \INFER[\text{\eqref{rule:Select2}, $\overline{v_2}$ occurs in $c_3$}]{[c_3] \SLASHF \ASSIGNCAT{v_1}{\TRUE} \SLASHF 
      \ASSIGNCAT{v_2}{\FALSE}}{%
      \PROJECT[3]{[c_3] \SLASHF [c_2]}{c_3}
      &
      \INFER[\text{\eqref{rule:Select1}, $v_1$ occurs in $c_2$}]{[c_2] \SLASHF \ASSIGNCAT{v_1}{\TRUE} \SLASHF 
        \ASSIGNCAT{v_2}{\FALSE}}{%
        \PROJECT[2]{[c_2] \SLASHF [c_1]}{c_2}
        &
        \INFER[\text{\eqref{rule:Select1}, $v_1$ occurs in $c_1$}]{[c_1] \SLASHF \ASSIGNCAT{v_1}{\TRUE} \SLASHF 
          \ASSIGNCAT{v_2}{\FALSE}}{%
          \PROJECT[1]{[c_1] \SLASHF [c_0]}{c_1}
          &
          \SUBTREE{[c_0] \SLASHF \ASSIGNCAT{v_1}{\TRUE} \SLASHF 
            \ASSIGNCAT{v_2}{\FALSE}}
        }
      }
    }
  \end{displaymath}
  \caption{Derivation fragment that verifies the assignment for the
    running example. The white triangle represents the derivation
    shown in Figure~\ref{fig:Guessing}.}
  \label{fig:Verifying}
\end{figure}

\paragraph{Lexicon Entries and Rules}

To implement the second part of the derivation, for each clause~$c_i$
we introduce a lexicon entry $c_i \LEXICON [c_i] \SLASHF [c_{i-1}]$.
Our rules make crucial use of variable restrictions.
To introduce them we define the following notational shorthands:
\begin{align*}
  \mathbf{1}_j &\equiv \SLASHF Y_1 \cdots \SLASHF Y_{j-1} \SLASHF \ASSIGNCAT{v_j}{\TRUE} \SLASHF Y_{j+1} \cdots \SLASHF Y_n\\
  \mathbf{0}_j &\equiv \SLASHF Y_1 \cdots \SLASHF Y_{j-1} \SLASHF \ASSIGNCAT{v_j}{\FALSE} \SLASHF Y_{j+1} \cdots \SLASHF Y_n
\end{align*}
Thus $\mathbf{1}_j$ is a sequence of $n$ slash"=variable pairs, except
that the $j$th variable has been replaced with the concrete (atomic)
category $\ASSIGNCAT{v_j}{\TRUE}$, and similarly for~$\mathbf{0}_j$.
With this notation, we include into~$G$ all rules that match one of
the following two schemata:
\begin{align}
  \RULE{X \SLASHF [c_{i-1}]}{[c_{i-1}] \mathbf{1}_j&}{X \mathbf{1}_j}
  && \text{if $v_j$ occurs in $c_i$}\label{rule:Select1}\\
  \RULE{X \SLASHF [c_{i-1}]}{[c_{i-1}] \mathbf{0}_j&}{X \mathbf{0}_j}
  && \text{if $\NEG{v_j}$ occurs in $c_i$}\label{rule:Select2}
\end{align}
For example, the two lowermost (when reading the tree from the root to
the leaves) compositions in Figure~\ref{fig:Verifying} are both
instances of schema \eqref{rule:Select1}, but their use is licensed by
two different variable"=clause matchings.

\paragraph{Derivational Ambiguity}

Similarly to what we noted for the first part
(Section~\ref{sec:NPHardnessPart1}), the derivation of this part can
proceed in several ways, because at each step we may be able to choose
more than one rule to satisfy a clause $c_i$.
For example, in the derivation in Figure~\ref{fig:Verifying}, instead
of using the rule of schema \eqref{rule:Select1} with witness ``$v_1$
occurs in $c_2$'' we could also have used the rule of schema
\eqref{rule:Select2} with witness ``$\overline{v_2}$ occurs in
$c_2$.''
However, also as before, there is no need to eliminate this derivational
ambiguity for the purposes of this reduction.

\subsubsection{Finalizing the Derivation}
\label{sec:NPHardnessPart3}

The third and final part of a derivation of~$w$ reduces the complex
category encoding the truth assignment to the distinguished category
of~$G$, which we define to be $[c_m]$, by a sequence of applications.
For the running example, this is illustrated in
Figure~\ref{fig:Finalizing}.

\begin{figure}
  \vskip-\abovedisplayskip
  \begin{displaymath}
    \INFER[\eqref{rule:Verifying1}]{[c_3]}{%
      \INFER[\eqref{rule:Verifying2}]{[c_3] \SLASHF \ASSIGNCAT{v_1}{\TRUE}}{%
        \SUBTREE{[c_3] \SLASHF \ASSIGNCAT{v_1}{\TRUE} \SLASHF 
          \ASSIGNCAT{v_2}{\FALSE}}
        &
        \PROJECT[1]{\ASSIGNCAT{v_2}{\FALSE}}{d_2}
       }
      &
      \PROJECT[2]{\ASSIGNCAT{v_1}{\TRUE}}{d_1}
    }
  \end{displaymath}
  \caption{Final derivation fragment for the running example.
    The white triangle represents the derivation shown in
    Figure~\ref{fig:Verifying}.}
  \label{fig:Finalizing}
\end{figure}

\paragraph{Lexicon Entries and Rules}

This part of the derivation requires two lexicon entries for each of
the auxiliary symbols: $d_j \LEXICON \ASSIGNCAT{v_j}{\TRUE}$ and
$d_j \LEXICON \ASSIGNCAT{v_j}{\FALSE}$.
The rules are:
\begin{align}
  \RULE{X \SLASHF \ASSIGNCAT{v_j}{\TRUE}}{\ASSIGNCAT{v_j}{\TRUE}&}{X} \label{rule:Verifying1}
  \\
  \RULE{X \SLASHF \ASSIGNCAT{v_j}{\FALSE}}{\ASSIGNCAT{v_j}{\FALSE}&}{X} \label{rule:Verifying2}
\end{align}

\subsubsection{Time Complexity}

We now analyze the time complexity of our reduction.
For a given clause $c_i$, let $\size{c_i}$ be the number of literals
in~$c_i$. 
We define $\size{\phi} = \sum_i \size{c_i}$.
The number of rules added in the first and the third part of the
construction of~$G$ is in $\order{n}$, and the size of each such rule
is bounded by a constant that does not depend on $\size{\phi}$.
For the second part of the construction of~$G$, for each clause $c_i$
we add a number of rules that is at most $\size{c_i}$, and possibly
less if there are repeated occurrences of some literal in~$c_i$.  
Thus the total number of rules added in this part is in
$\order{\size{\phi}}$.
Each such rule has size in $\order{n}$.
Putting everything together, and observing that $n$ is in
$\order{\size{\phi}}$, we see that the size of~$G$ is in
$\order{\size{\phi}^2}$.
It is not difficult to see then that our reduction can be carried out
in time polynomial in the size of~$\phi$.

\subsection{Membership in NP}
\label{sec:NPMembership}

We now turn to the membership part of Theorem~\ref{thm:NPCompleteness}:

\begin{lemma}\label{lem:NPMembership}
  The universal recognition problem for $\EMPTYSTRING$"-free \VWCCG\
  is in NP.
\end{lemma}

For the proof of this lemma, we provide a polynomial"-time
nondeterministic algorithm that accepts an $\EMPTYSTRING$"-free
\VWCCG\ $G$ and a string $w$ if and only if~$G$ can derive $w$.
We adopt the usual proof strategy where we first nondeterministically
guess a derivation tree for~$w$ and then verify that this tree is
valid.

\paragraph{Size of a Derivation Tree}

We need to argue that the total number of characters needed to encode
a derivation tree is polynomial in the combined size of~$G$ and~$w$.
Note that this involves both the tree structure itself and the lexicon
entries and categories at the nodes of the tree.
We start by observing that any derivation tree with $\ell$ leaf nodes
(labeled with lexicon entries) has exactly $\ell-1$ binary nodes
(labeled with categories).
Let $\tau$ by an arbitrary derivation tree for~$w$.
Since in $G$ there are no lexical categories for the empty string,
there is a one"-to"-one correspondence between the leaf nodes
of~$\tau$ and the symbols in~$w$, which implies that the number of
nodes of $\tau$ is exactly $2\size{w}-1$.

\paragraph{Maximal Size of a Category}

Now that we have bounded the number of nodes of~$\tau$, we will bound
the sizes of the categories that these nodes are labeled with.
Here, by the size of a category $X$, denoted by $|X|$, we simply mean
the number of characters needed to write down~$X$.
Consider an internal node $v$ of~$\tau$ and its associated category
$X$.
In order to state an upper bound for $\size{X}$, we distinguish two
cases: If $v$ is a unary node, then $\size{X}$ is bounded by the
largest size of a category in the lexicon of the grammar.
We denote this quantity by~$\lambda$.
If $v$ is a binary node, let $X = A \SLASH Y_1 \cdots \SLASH Y_q$,
with $A$ an atomic category.  
A rule of the grammar can increase the number of arguments of its
primary category by at most $d$, where $d$ is the maximum degree of a
rule in the grammar.
Let $\gamma$ be the maximum number of arguments in a category in the
lexicon.
Since no more than $\size{w}-1$ rules are used in~$\tau$, we conclude
that $q$ is bounded by $\gamma + d(\size{w}-1)$.  
By Lemma~3.1 in \citet{vijayshanker1994equivalence}, every argument
$Y_i$ in~$X$ must also occur as an argument in some category in the
lexicon of~$G$.
Thus the size of each argument of~$X$ is bounded by the largest size
of an argument appearing in a category in the lexicon, a quantity that
we denote by~$\alpha$. 
Putting everything together, we have that $\size{X}$ is bounded by
$1 + \alpha(\gamma + d(\size{w}-1))$.
From this it is not difficult to see that the overall space needed to
encode our derivation tree $\tau$ for~$w$ along with all of the
categories at its nodes is
$\order{(\lambda + \alpha \gamma) \size{w} + \alpha d \size{w}^2}$.
This is a polynomial in the combined size of the grammar and the input
string.

\paragraph{Nondeterministic Algorithm}

We can now provide our nondeterministic algorithm for testing whether
$G$ derives~$w$.
In a first step we write down a guess for a derivation tree $\tau$
for~$w$ based on the rules in $G$.
Given our space bound on~$\tau$, we can carry out this step in time
polynomial in the size of~$G$ and~$w$.
In a second step we visit each internal node $v$ of~$\tau$ and read
its associated category $X$.
If $v$ is a unary node, we check whether $X$ is a lexicon entry for
the word at $v$'s child.
If $v$ is a binary node, we check whether $X$ can be obtained by some
rule of the grammar applied to the categories at the two children
of~$v$.
We accept if every check is successful.
Even this second step can be carried out in time polynomial in the
size of~$G$ and~$w$.
This concludes the proof of Lemma~\ref{lem:NPMembership}.

\subsection{Discussion}
\label{sec:NPDiscussion}

In the previous sections we have shown that the universal recognition
problem for $\EMPTYSTRING$"-free \VWCCG\ is NP"-complete
(Theorem~\ref{thm:NPCompleteness}).
This result is in contrast with the fact that, for the weakly
equivalent \TAG\ formalism, the universal recognition problem can be
solved in polynomial time, and it naturally raises the question what
features of the \VWCCG\ formalism are the source of this additional
complexity.
We discuss this question on the basis of the reduction in our proof of
Lemma~\ref{lem:NPHardness}.
In this reduction we use a combination of three central features of
\VWCCG, listed below.
Dropping any of these features would break our reduction.

\paragraph{Lexical Ambiguity}

The first feature of \VWCCG\ that we exploit in our construction is
the ability to assign more than one category to some lexical items.
In part~1 of the reduction (Section~\ref{sec:NPHardnessPart1}), this
allows us to ``guess'' arbitrary truth assignments for the variables
in the clause~$\phi$.
However, the possibility to write grammars with lexical ambiguity is
an essential feature of all interesting formalisms for natural
language syntax, including also \TAG.
Therefore, at least in isolation, this feature does not seem to be
able to explain the complexity of the universal recognition problem
for \VWCCG.
Even if our goal was to design a new version of \VWCCG\ which can be
parsed in polynomial time in the size of the input grammar, we would
not seriously consider giving up lexical ambiguity.

\paragraph{Unbounded Composition}

The second feature of \VWCCG\ that we rely on is the availability of
composition rules without a constant (with respect to the full class
of grammars) bound on their degree.
This feature is crucial for our encoding of truth assignments.
In particular, without it we would not be able to percolate
arbitrarily large truth assignments through derivation trees; our
construction would work only for formulas with a bounded number of
variables.

Unbounded composition has previously been discussed primarily in the
context of generative power.
\citet{weir1988combinatory} show that allowing unrestricted use of
\emph{arbitrarily} many composition rules leads to a version of
\VWCCG\ that is more powerful than the one considered here, where
every grammar must restrict itself to a finite set of such rules.
Other authors have suggested to put explicit (low) bounds on the
maximal degree of composition.
From a purely formal point of view, a bound as low as $d \leq 2$ may
suffice: \citet{weir1988combinatory} show how every \VWCCG\ grammar
can be converted into a weakly equivalent Linear Index Grammar (LIG)
\citep{gazdar1987applicability}, and how every \TAG\ can be converted
into a weakly equivalent \VWCCG\ whose composition rules all have
degree~$2$.
Together with the weak equivalence of \LIG\ and \TAG\
\citep{vijayshanker1985some}, this shows that the subclass of \VWCCG\
grammars with degree of composition at most~2 can still generate the
full class of languages.\footnote{Note however that the construction
  of \citet{weir1988combinatory} does not produce $\EMPTYSTRING$-free
  grammars.}
For any degree"-restricted subclass of grammars, our proof would
break, which means that it may be possible (though not obvious) to
devise a polynomial"-time algorithm for the universal recognition
problem.
We will discuss unbounded composition further in
Section~\ref{sec:UnboundedComposition}.

\paragraph{Rule Restrictions}

The third feature of \VWCCG\ that we exploit is its ability to put
grammar"-specific restrictions on combinatory rules.
In particular, in part~2 of our construction
(Section~\ref{sec:NPHardnessPart2}), we use rules whose secondary
input categories contain a mix of variables and concrete categories,
such as
\begin{align*}
  \RULE{X \SLASHF [c_{i-1}]}{[c_{i-1}] \mathbf{1}_j&}{X \mathbf{1}_j}
  && \text{if $v_j$ occurs in $c_i$} \tag{\ref{rule:Select1}}
\end{align*}
Like the availability of composition rules of unbounded degree, the
ability to use rule restrictions seems to be a very powerful feature,
and one that perhaps most clearly sets \VWCCG\ apart from \TAG.
Moreover, as already mentioned, rule restrictions also play a crucial
role with respect to weak generative capacity
\citep{kuhlmann2015lexicalization}.

Note that we could replace rules of the form~\eqref{rule:Select1} with
rules without variables; but then, for fixed values of~$i$ and $j$
and, say, for the assignment $\ASSIGNCAT{v_j}{\TRUE}$, we would have
to include into the grammar all of the $2^{n-1}$ rules of the form
\begin{displaymath}
  \RULE{X \SLASHF [c_{i-1}]}{[c_{i-1}] \SLASHF A_1 \cdots \SLASHF A_{j-1} 
    \SLASHF \ASSIGNCAT{v_j}{\TRUE} \SLASHF A_{j+1} \cdots \SLASHF A_n}{X \SLASHF 
    A_1 \cdots \SLASHF A_n}
\end{displaymath}
where each $A_h$ is a concrete atomic category of the form
$\ASSIGNCAT{v_h}{\TRUE}$ or $\ASSIGNCAT{v_h}{\FALSE}$.
This would break our proof because reductions must use polynomial time
(and space).
Note also that what is crucial here is not the use of either variables
or concrete categories in a rule's secondary input; rather, it is the
combination of the two that allows us to check clauses against truth
assignments.

\section{Complexity With Categories for the Empty String}
\label{sec:EXPTIMECompleteness}

In this section we investigate the computational complexity of the
universal recognition problem for unrestricted \VWCCG, where one is
allowed to assign lexicon entries even to the empty string.
We show the following:
\begin{theorem}\label{thm:EXPTIMECompleteness}
  The universal recognition problem for unrestricted \VWCCG\ is
  \EXPTIME"-complete.
\end{theorem}
The proof of this theorem is more involved than the proof of the
NP"-completeness result in Section~\ref{sec:NPCompleteness}.
We start in Section~\ref{sec:ATM} by introducing \emph{alternating
  Turing machines} \citep{chandra1981alternation}, which provide the
computational framework for our proof.
The use of alternating Turing machines instead of ordinary
deterministic or nondeterministic Turing machines is crucial here: In
order to simulate the computations of a Turing machine by a \CCG\
grammar in a natural way, we need to restrict the machine to use only
polynomial space.
However, if we used standard Turing machines with this space
restriction, then we would only be able to prove PSPACE"-hardness, a
weaker result than the EXPTIME"-completeness that we obtain from our
proof.
The hardness part of this proof is presented in
Section~\ref{sec:EXPTIMEHardness}, and the membership part in
Section~\ref{sec:EXPTIMEMembership}.
We finally discuss our result in Section~\ref{sec:EXPTIMEDiscussion}.


\subsection{Alternating Turing Machines}
\label{sec:ATM}

The \TERMF{alternating Turing machine} (ATM;
\citealp{chandra1981alternation}) is a generalization of the
well"-known nondeterministic Turing machine in which there are two
types of states: \TERMF{existential} states and \TERMF{universal}
states.
When the machine is in an existential state, it accepts the input if
there is at least one transition that eventually leads to an accepting
state.
In contrast, when the machine is in a universal state, it accepts only
if \emph{every} possible transition eventually leads to an accepting
state.
A nondeterministic Turing machine can be viewed an alternating Turing
machine with no universal states.

As already mentioned, for our proof we use ATMs working in polynomial
space, which means that the length of the tape is bounded by a
polynomial in the length of the input.
This resource"-restricted model is well"-known in the literature, and
it exactly characterizes the class of all decision problems that are
solvable by a \emph{deterministic} Turing machine (i.e.\ a Turing
machine where there is at most one possible transition given a state
and a tape symbol) working in \emph{exponential} time
\citep{chandra1981alternation}.
This is the complexity class EXPTIME.

To simplify the notation and some of our proofs, we use ATMs that
operate on a \emph{circular tape}, and can only move their head to the
right.
The same model has previously been used by, among others,
\citet{jez2011complexity}.
It is not hard to see that, as long as we work under the restriction
to polynomial space, every move to the left in the standard model can
be simulated by a (polynomial) number of moves to the right in the
circular tape model.
Thus, even ATMs with a polynomially bounded circular tapes precisely
characterize EXPTIME.

\paragraph{Formal Definition}

Formally, an \TERM{alternating Turing machine} is a structure
\begin{displaymath}
  \myTM = (Q, \Sigma, \delta, q_0, g)
\end{displaymath}
where:
\begin{inparablank}
\item $Q$ is a finite set of \TERM{states};
\item $\Sigma$ is an alphabet of \TERM{tape symbols}, which we assume
  includes the special blank symbol~$\#$;
\item $\delta \subseteq (Q \times \Sigma) \times (Q \times \Sigma)$ is
  the \TERM{transition relation};
\item $q_0 \in Q$ is the \TERMF{initial state};
\item and
  $g\mathpunct{:}\ Q \to \SET{{\exists},{ \forall}, \ACCEPT, \REJECT}$
  is a function that assigns a type to each state.
\end{inparablank}
The four different types for a state are \TERMF{existential}
($\exists$), \TERMF{universal} ($\forall$), \TERMF{accepting}
$(\ACCEPT)$, and \TERMF{rejecting} $(\REJECT)$; their semantics will
become clear below.

We denote transitions in~$\delta$ as $(q, a) \to (q', a')$.
Transitions are subject to the restriction that the state to the left
of the arrow must be either existential or universal.
This means that no transition is possible out of an accepting or a
rejecting state; when an ATM reaches such a state, it necessarily
stops.
To simplify the proof, we also require that for every universal state
$q$ and tape symbol $a$, there are exactly two transitions with
left"-hand side $(q, a)$.
This is without loss of generality: If a machine does not already have
this property, then one can construct (in polynomial time) an
equivalent polynomial"-space ATM with circular tape satisfying it; a
similar construction for general ATMs is sketched by
\citet[Theorem~8.2]{papadimitriou1994computational}.


\paragraph{Configurations}

Let $w \in \Sigma^*$ be an input string for~$w$, and let
$n = \size{w}$ and $m = p_{\myTM}(\size{w})$, where $p_{\myTM}$ is the
machine"-specific polynomial that defines the maximal tape length.
A \TERM{configuration} of~$\myTM$ relative to~$w$ is a pair
$c = (q, \alpha)$, where $q \in Q$ is some state and
$\alpha \in \Sigma^*$ is a sequence of tape symbols with length
$\size{\alpha} = m$.
The intended interpretation of~$c$ is that the current state
of~$\myTM$ is $q$, the content of the circular tape is represented
by~$\alpha$, and the tape head is positioned to read the first symbol
of~$\alpha$.
In particular, the \TERMF{initial} configuration of~$\myTM$ for~$w$,
denoted by $I_M(w)$, takes the form $I_M(w) = (q_0, w\#^{m-n})$,
meaning that, at the start of the computation, the machine is in the
initial state, the tape consists of the $n$ symbols of the input
string $w$ followed by $m-n$ blanks, and the tape head is positioned
to read the first symbol of~$w$.
A configuration~$c$ is called \TERMF{existential}, \TERMF{universal},
\TERMF{accepting} or \TERMF{rejecting}, based on the type of its
state~$q$.

\paragraph{Successors}

Let $t = (q, a) \to (q', a')$ be a transition.
The intended interpretation of~$t$ is that if $\myTM$ is in state $q$
and reads tape symbol $a$, then overwrites $a$ with $a'$, moves its
tape head one cell to the right (which is always possible because the
tape is circular), and continues the computation in state~$q'$.
Formally, let $c = (q, a \alpha)$ be a configuration of~$\myTM$.
The \TERM{successor} of~$c$ with respect to~$t$, denoted by $t(c)$, is
the configuration $c' = (q', \alpha a')$, where the string $\alpha a'$
encodes the fact that the symbol $a$ has been overwritten with~$a'$
and the circular tape has been rotated one position to the right, so
that the head now is posititioned to read the first symbol
of~$\alpha$.
Note that, due to our restrictions on the transition relation, a
universal configuration has exactly two successors.

\paragraph{Acceptance}

We first discuss acceptance in ordinary nondeterministic Turing
machines. As usual, a single machine configuration $c$ may lead (in
one step) to a number of successor configurations $c_1,\dots,c_k$.
Acceptance is recursively defined such that $c$ leads to acceptance if
and only if at least one of $c_1,\dots,c_k$ leads to acceptance.
One may view this as an existential condition on the successor
configurations.
In an alternating Turing machine, a configuration may be either
existential or universal; in the universal case, $c$ leads to
acceptance if and only if \emph{every} successor $c_1,\dots,c_k$ leads
to acceptance.
To make this formal, we represent computations of~$\myTM$ as trees
whose nodes are labeled with configurations of~$\myTM$, and whose
edges reflect the ``successor of''"-relation between configurations.
Formally, the set of \TERM{accepting computations} is defined
recursively as follows (recall our definition of tree terms in
Section~\ref{ssec:grammar}):
\begin{itemize}
\item\relax Every accepting configuration $c$ forms a one"-node
  accepting computation.\medskip
\item\relax Let $c$ be an existential configuration and let $\gamma$
  be an accepting computation whose root node is labeled with some
  successor of~$c$.
  Then $c(\gamma)$ is an accepting computation.\medskip
\item\relax Let $c$ be a universal configuration and let $\gamma_1$,
  $\gamma_2$ be accepting computations whose root nodes are labeled
  with the two successors of~$c$.
  Then $c(\gamma_1, \gamma_2)$ is an accepting computation.
\end{itemize}

\noindent A sample accepting computation is shown in
Figure~\ref{fig:computation}.
A machine $\myTM$ \TERM{accepts} a string $w$ if there exists an
accepting computation $\gamma$ whose root node is labeled with the
initial configuration $I_M(w)$.
The set of all strings that are accepted by $\myTM$ is denoted by
$L(\myTM)$.

\paragraph{Characterization of EXPTIME}

As already mentioned, the reason that we are interested in
polynomial"-space alternating Turing machines is that they exactly
characterize the class of decision problems solvable in exponential
time.
This is expressed by the following lemma, which is basically
Corollary~3.6 of \citet{chandra1981alternation}.
\begin{lemma}\label{polyatmhard}
  The following decision problem is EXPTIME"-complete: Given a
  polynomial"-space alternating Turing machine $\myTM$ and a string
  $w$, is $w \in L(\myTM)$?
\end{lemma}
Since a polynomial"-space circular"-tape ATM can simulate any
polynomial"-space ATM at no additional asymptotic space cost, we
conclude that Lemma~\ref{polyatmhard} also holds for polynomial"-space
circular"-tape ATMs.
In the following we therefore use Lemma~\ref{polyatmhard} as referring
to polynomial space circular tape ATMs. 

\begin{figure}
  \begin{center}
    \begin{tikzpicture}
      \node(c1) at (1, 3) {$c_1$};
      \node[right=0mm of c1, anchor=west] {$(\exists)$};
      \node(c2) at (1, 2) {$c_2$};
      \node[right=0mm of c2, anchor=west] {$(\forall)$};
      \node(c3) at (0, 1) {$c_3$};
      \node[left=0mm of c3, anchor=east] {$(\ACCEPT)$};
      \node(c4) at (2, 1) {$c_4$};
      \node[right=0mm of c4, anchor=west] {$(\exists)$};
      \node(c5) at (2, 0) {$c_5$};
      \node[right=0mm of c5, anchor=west] {$(\ACCEPT)$};
      \draw (c1) -- (c2);
      \draw (c2) -- (c3);
      \draw (c2) -- (c4);
      \draw (c4) -- (c5);
    \end{tikzpicture}
  \end{center}
  \caption{An accepting run.
    State types for each configuration $c_i$ are indicated in
    parentheses.}
  \label{fig:computation}
\end{figure}

\subsection{EXPTIME"-Hardness}
\label{sec:EXPTIMEHardness}

Let $\myTM$ be a polynomial"-space circular"-tape ATM and let $w$ be
an input string for~$\myTM$.
In this section we show how to construct, in polynomial time and
space, a \VWCCG\ grammar $G$ such that $L(G) = \SET{\ep}$ if
$w \in L(\myTM)$, and $L(G) = \emptyset$ if $w \not \in L(\myTM)$.
This means that we can test the condition $w \in L(\myTM)$ by checking
whether $G$ generates the empty string.
When combined with Lemma~\ref{polyatmhard}, this reduction proves the
hardness part of Theorem~\ref{thm:EXPTIMECompleteness}:

\begin{lemma}\label{lem:EXPTIMEHardness}
  The universal recognition problem for unrestricted \VWCCG\ is
  \EXPTIME"-hard.
\end{lemma}

For the remainder of this section, we fix a polynomial"-space
circular"-tape ATM $\myTM = (Q, \Sigma, \delta, q_0, g)$ and an input
string $w \in \Sigma^*$.
Let $p_{\myTM}$ be the polynomial that bounds the length of the tape
of~$\myTM$, and let $m = p_{\myTM}(\size{w})$.

\paragraph{Basic Idea}

The basic idea behind our construction is straightforward: We will set
up things in such a way that the derivations of~$G$ correspond to
accepting computations of~$\myTM$ for~$w$.
To illustrate this idea, Figure~\ref{fig:derivation} shows the
schematic structure of a derivation that corresponds to the accepting
computation in Figure~\ref{fig:computation}.
Note that in order to make the correspondence more evident, contrary
to our previous convention, we now draw the derivation with the root
node at the top.
We see that the derivation is composed of a number of smaller
fragments (drawn as triangles).
With the exception of the fragment at the top of the tree (which we
need for technical reasons), there is one fragment per node of the
accepting computation.
Each fragment is labeled with a reference to the subsection of this
article that describes how we set up the grammar $G$ to derive that
fragment.

\begin{figure}
  \begin{center}
    \newcommand*\mytriangle[3]{%
      \draw (#1) node[fill=white] {#2} -- ++(1,-2) -- ++(-2,0) -- (#1);
      \node (#3) at ($(#1) + (0,-1.1)$) {};
    }
    \newcommand*\myshadedtriangle[3]{%
      \draw[fill=yellow!60] (#1) node[fill=white] {#2} -- ++(1,-2) -- ++(-2,0) -- (#1);
      \node (#3) at ($(#1) + (0,-1.1)$) {};
    }
    
    \begin{tikzpicture}[y=0.618cm,mystyle/.style={every text node part/.style={align=left},font=\footnotesize}]
      \myshadedtriangle{0,8}{$[\CMD{init}]$}{a} 
      \mytriangle{0,6}{$E(c_1)$}{b} 
      \begin{scope}[x=2.5cm]
        \mytriangle{0,4}{$E(c_2)$}{c} 
      \end{scope}
      \mytriangle{-1.25,2}{$E(c_3)$}{d} 
      \mytriangle{1.25,2}{$E(c_4)$}{e} 
      \mytriangle{1.25,0}{$E(c_5)$}{f} 
      \node[anchor=west,mystyle] (alabel) at ($(a) + (3.5, 0)$) {initial configuration \\ (\ref{fragment:InitialConfiguration})};
      \draw[dashed] (a) -- (alabel);
      \node[anchor=west,mystyle] (blabel) at ($(b) + (3.5, 0)$) {transition out of an \\ existential configuration \\ (\ref{fragment:ExistentialConfigurations})};
      \draw[dashed] (b) -- (blabel);
      \node[anchor=west,mystyle] (clabel) at ($(c) + (3.5, 0)$) {transition out of a \\ universal configuration \\ (\ref{fragment:UniversalConfigurations})};
      \draw[dashed] (c) -- (clabel);
      \node[anchor=east,mystyle,every text node part/.style={align=right}] (dlabel) at ($(d) - (2.25, 0)$) {accepting configuration \\ (\ref{fragment:AcceptingConfigurations})};
      \draw[dashed] (d) -- (dlabel);
      \node[anchor=west,mystyle] (elabel) at ($(e) + (2.25, 0)$) {transition out of an \\ existential configuration \\ (\ref{fragment:ExistentialConfigurations})};
      \draw[dashed] (e) -- (elabel);
      \node[anchor=west,mystyle] (flabel) at ($(f) + (2.25, 0)$) {accepting configuration \\ (\ref{fragment:AcceptingConfigurations})};
      \draw[dashed] (f) -- (flabel);
    \end{tikzpicture}
  \end{center}
  \caption{Schematic structure of the derivation in $G$ that
    corresponds to the accepting computation in
    Figure~\ref{fig:computation}. The derivation is displayed with the
    root node at the top to make the correspondence with
    Figure~\ref{fig:computation} more evident. Each $E(c_i)$ is the
    category that encodes the configuration~$c_i$.}
  \label{fig:derivation}
\end{figure}

One way to view our construction is that it establishes a
structure"-preserving map from accepting computations of~$\myTM$ to
derivations of~$G$.
This map replaces each configuration $c$ in an accepting computation
by a fragment, and continues the transformation at the subtrees
below~$c$.
A fragment is like a small derivation tree, except that one or two of
its leaf nodes may be labeled with (possibly complex) categories
instead of lexicon entries.
These nodes, which we refer to as the \TERM{distinguished} leaf nodes
of the fragment, serve as slots at which the fragments that result
from the recursive transformation of the subtrees can be substituted.
The root node of a fragment is labeled with a category that encodes
the configuration $c$ that the fragment replaces; we denote this
category by $E(c)$.
More specifically, the shape of the fragment depends on the type
of~$c$:

\begin{itemize}

\item For every accepting configuration $c$, the grammar derives a
  fragment with no distinguished leaf nodes.
  The category at the root node of this fragment is $E(c)$.
  The lexicon entries and rules required to derive the fragment are
  described in Section~\ref{fragment:AcceptingConfigurations}.
  \medskip

\item For every existential configuration $c$ and for every transition
  $t$ that can be applied to~$c$, the grammar derives a fragment with
  a single distinguished leaf node.
  The category at the root node of this fragment is $E(c)$, and the
  category at the distinguished leaf node is $E(t(c))$, the encoding
  of the successor of~$c$ under~$t$.
  (Section~\ref{fragment:ExistentialConfigurations})\medskip

\item For every universal configuration $c$, the grammar derives a
  fragment with two distinguished leaf nodes.
  The category at the root node of this fragment is $E(c)$, and the
  categories at the distinguished leaf nodes are $E(t_1(c))$ and
  $E(t_2(c))$, the encodings of the two successors of~$c$.
  (Section~\ref{fragment:UniversalConfigurations})\medskip

\item Finally, for the initial configuration $I_M(w)$, the grammar
  derives a fragment with a single distinguished leaf node.
  The category at the root node of this fragment is the distinguished
  category of~$G$, and the category at the distinguished leaf node is
  $E(I_M(w))$.
  This is the highlighted fragment in Figure~\ref{fig:derivation}.
  (Section~\ref{fragment:InitialConfiguration})

\end{itemize}

\noindent Those leaf nodes of a fragment that are not distinguished
leaf nodes will always be labeled with lexicon entries for the empty
string, that is, entries of the form $\EMPTYSTRING \LEXICON X$.
Because of this, the only string that our grammar may accept is the
empty string.
As we will make sure that all (and only) the accepting computations
of~$\myTM$ over~$w$ receive corresponding derivations in~$G$, this
amounts to saying that $G$ has at least one derivation if and only if
$w \in L(\myTM)$.
This implies that we do not need a distinguished ``reject'' lexical
category or any other mechanism that takes care of the case when the
Turing machine rejects its input.

\paragraph{Technical Remark}

Before we continue, we would like to make a technical remark that may
make the following construction easier to understand.
In the proof of Lemma~\ref{lem:NPHardness}, we constructed a grammar
that produced derivations simulating a process of guessing and
verifying a variable assignment for an instance of SAT.
One feature of the construction is that this process has a purely
linear (albeit nondeterministic) structure, which is reflected in the
fact that the derivation trees produced by the grammar are essentially
unary"-branching.
For such trees, it does not make much of a difference whether we read
them bottom"=up (from the leaves to the root) or top"=down (from the
root to the leaves), and in our construction we simply adopted the
former perspective, which is the conventional one for \CCG.

In this proof, because of the branching nature of the computations
of~$\myTM$, the derivation trees of the grammar~$G$ will no longer be
unary"-branching; and because the branching in an accepting
computation of~$\myTM$ occurs on the paths from the initial
configuration to the accepting configurations, the derivation trees of
the grammar~$G$ need to have the encoding of the initial configuration
at the root and the encodings of the accepting configurations at the
leaves.
This will require us to change perspective and read the derivation
trees top"=down---and consequently the rules of~$G$ from the output
category to the input categories.
This is the reverse of what is conventional for \CCG.

\subsubsection{Encoding Configurations}

We start the presentation of the construction of~$G$ by explaining how
we encode configurations of~$\myTM$ as categories.
Let $c = (q, a_1 \cdots, a_m)$ be a configuration of~$\myTM$.
We encode this configuration by a category
\begin{displaymath}
  E(c) = \CONTROL{q} \SLASHF [a_1] \cdots \SLASHF [a_m]
\end{displaymath}
where we follow the same convention as in Section~\ref{sec:NPHardness}
and use square brackets to represent atomic categories.
Note that in this encoding, the target of $E(c)$ is an atomic category
representing the current state, while the arguments of $E(c)$
represent the circular tape, with the innermost argument corresponding
to the symbol under the tape head.
With this representation, the encoding of the successor of the
configuration $c$ under a transition $t = (q, a_1) \to (q', a')$ can
be written as
\begin{displaymath}
  E(t(c)) = \CONTROL{q'} \SLASHF [a_2] \cdots \SLASHF [a_m] \SLASHF [a']
\end{displaymath}

\subsubsection{Initial Configuration}
\label{fragment:InitialConfiguration}

We now present the derivation fragment for the initial configuration
of~$\myTM$ for~$w$.
Let $c_I = I_\myTM(w)$.
To give a concrete example, suppose that this configuration takes the
form $c_I = (q_0, a b)$.
Then the corresponding fragment looks as in Figure~\ref{fig:Initial}.
The category at the root node is the distinguished category
$\CONTROL{\CMD{init}}$.
The derivation starts by nondeterministically pushing symbols to the
tape stack of the categories along the path (highlighted in
Figure~\ref{fig:Initial}) from the root to the distinguished node.
This is done through rules of type~\eqref{rule:Init1} below. 
In a last step \eqref{rule:Init2}, the derivation checks whether the
tape stack matches the initial tape content $a b$, and simultaneously
changes the target from $\CONTROL{\CMD{init}}$ to $\CONTROL{q_0}$.
After this, the category at the distinguished leaf of the fragment is
$E(c_I)$.
We will see the idea of ``nondeterministic guessing followed by
verification'' once again in
Section~\ref{fragment:UniversalConfigurations}, where it will be used
for making copies of the tape stack.
The reader may rightfully wonder whether there are more direct methods
for performing such straightforward manipulations of the tape stack.
Unfortunately, we have not been able to define such methods within the
VW-CCG formalism.

\begin{figure}
  \vskip-\abovedisplayskip
  \begin{displaymath}
    \INFER[\eqref{rule:Init1}]{\HIGHLIGHT{\CONTROL{\CMD{init}}}}{%
      \INFER[\eqref{rule:Init1}]{\HIGHLIGHT{\CONTROL{\CMD{init}} \SLASHF [a]}}{%
        \INFER[\eqref{rule:Init2}]{\HIGHLIGHT{\CONTROL{\CMD{init}} \SLASHF [a] \SLASHF [b]}}{%
          \PROJECT{\CONTROL{\CMD{init}} \SLASHF \CONTROL{q_0}}{\EMPTYSTRING}
          &
          \SUBTREE{E(c_I) = \HIGHLIGHT{\CONTROL{q_0} \SLASHF [a] \SLASHF [b]}}
        }
        &
        \PROJECT[2]{[b]}{\EMPTYSTRING}
      }
      &
      \PROJECT[3]{[a]}{\EMPTYSTRING}
    }
  \end{displaymath}
  \caption{Derivation fragment for an initial configuration
    $c_I = (q_0, a b)$.
    Note how the symbols of $c_I$ are collected in several steps along
    the highlighted path.}
  \label{fig:Initial}
\end{figure}

\paragraph{Lexicon Entries and Rules}

More generally now, assume that the initial configuration for~$M$
on~$w$ is $c_I = (q_0, a_1 \cdots a_m)$, where $w = a_1 \cdots a_n$
and $a_h = \#$ for each $h$ with $n < h \leq m$.
To support fragments as the one in Figure~\ref{fig:Initial}, we
introduce lexicon entries
$\EMPTYSTRING \LEXICON \CONTROL{\CMD{init}} \SLASHF \CONTROL{q_0}$ and
$\EMPTYSTRING \LEXICON [a]$, where $a \in \Sigma$ is any tape symbol.
We also introduce the following rules:
\begin{align}
  \RULE{\CONTROL{\CMD{init}} \STACK \SLASHF [a]}{[a]&}{\CONTROL{\CMD{init}} \STACK}\label{rule:Init1}\\
  \RULE{\CONTROL{\CMD{init}} \STACK \SLASHF \CONTROL{q_0}}{\CONTROL{q_0} \SLASHF [a_1] \cdots \SLASHF [a_m]&}{\CONTROL{\CMD{init}} \STACK \SLASHF [a_1] \cdots \SLASHF [a_m]}\label{rule:Init2}
\end{align}
The $\STACK$ symbol is, as usual, a variable for the part of the
category stack below the topmost stack element.
A rule of the form~\eqref{rule:Init1} allows the application of a
category with target $\CONTROL{\CMD{init}}$ to any atomic category
$[a]$ representing a tape symbol; this implements the nondeterministic
pushing to the tape stack that we introduced above.
A rule of the form~\eqref{rule:Init2} is a composition rule of degree
$m$ that restricts the target of the primary input category to the
distinguished category $\CONTROL{\CMD{init}}$, and the secondary input
to the category $E(c_I)$.
This implements the check in the final step of the fragment---if the
category at the distinguished leaf does \emph{not} encode the initial
configuration at this point, then the derivation will reach a dead
end.

\paragraph{Computational Complexity}

We now deal with the computational resources required by this step of
the construction.
Each of the lexicon entries above is size"-bounded by a constant that
does not depend on~$\size{M}$ or $\size{w}$.
This size bound also holds for each rule of the
form~\eqref{rule:Init1}.
The size of a rule of the form~\eqref{rule:Init2} is in $\order{m}$.
We can then construct and store each lexical entry and each rule with
time (and space) in $\order{m}$.
Furthermore, the total number of lexicon entries and rules added to
the grammar in this step is in $\order{\size{\Sigma}}$.
We thus conclude that this step of the construction can be carried out
in time (and space) polynomial in $\size{\myTM}$ and $\size{w}$.

\subsubsection{Accepting Configurations}
\label{fragment:AcceptingConfigurations}

Next we present lexicon entries and rules needed to terminate
derivations of the accepting configurations of~$\myTM$.
To give a concrete example, suppose that $c = (q, a b)$ is accepting,
and that the grammar has already derived the category $E(c)$.
Then the grammar also derives the fragment shown in
Figure~\ref{fig:Accepting}.
Following the highlighted path from the root to the leaf, the
derivation first checks whether $E(c)$ indeed encodes an accepting
configuration, and then changes the target to a special atomic
category $[\CMD{accept}]$ \eqref{rule:Accept1}.
After this, the fragment empties the tape stack, using derivations
similar to those that we used to assemble the truth assignment in
Section~\ref{sec:NPHardnessPart1} (see Figure~\ref{fig:Guessing}).

\begin{figure}
  \vskip-\abovedisplayskip
  \begin{displaymath}
    \INFER[\eqref{rule:Accept1}]{E(c) = \HIGHLIGHT{\CONTROL{q} \SLASHF [a] \SLASHF [b]}}{%
      \PROJECT[4]{\CONTROL{q} \SLASHF \CONTROL{\CMD{accept}}}{\EMPTYSTRING}
      &
      \INFER[\eqref{rule:Accept2}]{\HIGHLIGHT{\CONTROL{\CMD{accept}} \SLASHF [a] \SLASHF [b]}}{%
        \INFER[\eqref{rule:Accept3}]{\HIGHLIGHT{\CONTROL{\CMD{accept}} \SLASHF [a] \SLASHF [b] \SLASHF [\TECH]}}{%
          \INFER[\eqref{rule:Accept3}]{\HIGHLIGHT{\CONTROL{\CMD{accept}} \SLASHF [a] \SLASHF [\TECH]}}{%
            \PROJECT{\HIGHLIGHT{\CONTROL{\CMD{accept}} \SLASHF [\TECH]}}{\EMPTYSTRING}
            &
            \PROJECT{[\TECH] \SLASHF [a] \SLASHF [\TECH]}{\EMPTYSTRING}
          }
          &
          \PROJECT[2]{[\TECH] \SLASHF [b] \SLASHF [\TECH]}{\EMPTYSTRING}
        }
        &
        \PROJECT[3]{[\TECH]}{\EMPTYSTRING}
      }
    }
  \end{displaymath}
  \caption{Derivation fragment for an accepting configuration
    $c = (q, a b)$.}
  \label{fig:Accepting}
\end{figure}

\paragraph{Lexicon Entries and Rules}

Let $q \in Q$ with $g(q) = \ACCEPT$, and let $a \in \Sigma$.
We introduce the following lexicon entries:
\begin{displaymath}
  \EMPTYSTRING \LEXICON \CONTROL{q} \SLASHF \CONTROL{\CMD{accept}}
  \qquad
  \EMPTYSTRING \LEXICON \CONTROL{\CMD{accept}} \SLASHF [\TECH]
  \qquad
  \EMPTYSTRING \LEXICON [\TECH] \SLASHF [a] \SLASHF [\TECH]
  \qquad
  \EMPTYSTRING \LEXICON [\TECH]
\end{displaymath}
We also introduce the following rules:
\begin{align}
  \RULE{\CONTROL{q} \STACK \SLASHF \CONTROL{\CMD{accept}}}{\CONTROL{\CMD{accept}} \SLASHF X_1 \cdots \SLASHF X_m&}{\CONTROL{q} \STACK \SLASHF X_1 \cdots \SLASHF X_m}\label{rule:Accept1}\\
  \RULE{\CONTROL{\CMD{accept}} \STACK \SLASHF [\TECH]}{[\TECH]&}{\CONTROL{\CMD{accept}} \STACK}\label{rule:Accept2}\\
  \RULE{\CONTROL{\CMD{accept}} \STACK \SLASHF [\TECH]}{[\TECH] \SLASHF [a] \SLASHF [\TECH]&}{\CONTROL{\CMD{accept}} \STACK \SLASHF [a] \SLASHF [\TECH]}\label{rule:Accept3}
\end{align}
A rule of the form~\eqref{rule:Accept1} is a composition rule of
degree $m$ that restricts the target of its primary input to an
accepting state; this ensures that only categories encoding accepting
configurations can yield subderivations of the form shown in
Figure~\ref{fig:Accepting}.
The derivation will either rewrite the configuration (for existential
and universal configurations---the details will be given in
Sections~\ref{fragment:ExistentialConfigurations} and
\ref{fragment:UniversalConfigurations}, respectively), or else will
reach a dead end (for rejecting configurations).
The only rules that can be used after a rule of the form
\eqref{rule:Accept1} are rules of the forms~\eqref{rule:Accept2}
and~\eqref{rule:Accept3}, which jointly implement the emptying of the
tape stack that we described above.

\paragraph{Computational Complexity}

Each of the lexicon entries above is size"-bounded by a constant that
does not depend on $\size{M}$ and $\size{w}$.
This size bound also holds for rules of the forms~\eqref{rule:Accept2}
and~\eqref{rule:Accept3}.
The sizes of rules of the form~\eqref{rule:Accept1} are in
$\order{m}$.
The number of lexicon entries and rules that we add to~$G$ in this
step is in $\order{\size{\Sigma}}$.

\subsubsection{Transitions Out of Existential Configurations}
\label{fragment:ExistentialConfigurations}

We now turn to the fragments that simulate transitions out of
existential configurations.
Figure~\ref{fig:ExistentialState} shows how such a fragment looks like
for a configuration $c = (q, a b)$ and a transition
$t = (q, a) \to (q', a')$.
The derivation starts by checking whether the category at the root
node indeed encodes an existential configuration, and then changes the
target to the transition"-specific category $\CONTROL{t}$
\eqref{rule:Existential1}.
The derivation then extends the tape stack by the new symbol $a'$
\eqref{rule:Existential2}.
In a last step it simultaneously discards the category for the
previous tape symbol $a$ and changes the target to $\CONTROL{q'}$
\eqref{rule:Existential3}.
After this, the category at the distinguished leaf encodes the
configuration $t(c) = (q', b a')$.
We remind the reader that an ATM accepts in an existential
configuration if and only if there is at least one transition that
leads to an accepting configuration.
In the grammar $G$, this corresponds to the fact that there exists an
applicable rule that leads to acceptance.
Therefore, we do not need to explicitly simulate all possible
transitions.
For universal configurations (which we will consider in
Section~\ref{fragment:UniversalConfigurations}), the situation is
different, which will necessitate a more involved construction.

\begin{figure}
  \vskip-\abovedisplayskip
  \begin{displaymath}
    \INFER[\eqref{rule:Existential1}]{E(c) = \HIGHLIGHT{\CONTROL{q} \SLASHF [a] \SLASHF [b]}}{%
      \PROJECT[3]{\CONTROL{q} \SLASHF \CONTROL{t}}{\EMPTYSTRING}
      &
      \INFER[\eqref{rule:Existential2}]{\HIGHLIGHT{\CONTROL{t} \SLASHF [a] \SLASHF [b]}}{%
        \INFER[\eqref{rule:Existential3}]{\HIGHLIGHT{\CONTROL{t} \SLASHF [a] \SLASHF [b] \SLASHF [a']}}{%
          \PROJECT{\CONTROL{t} \SLASHF [a] \SLASHF [q']}{\EMPTYSTRING} &
          \SUBTREE{E(t(c)) = \HIGHLIGHT{\CONTROL{q'} \SLASHF [b] \SLASHF [a']}}
        }
        &
        \PROJECT[2]{[a']}{\EMPTYSTRING}
      }
    }
  \end{displaymath}
  \caption{Fragment for a transition $t = (q, a) \to (q', a')$ out of
    an existential configuration $c = (q, a b)$.}
  \label{fig:ExistentialState}
\end{figure}

\paragraph{Lexicon Entries and Rules}

Let $q \in Q$ be any existential state, and let
$t = (q, a) \to (q', a')$ be any transition out of~$q$.
We introduce the following new lexicon entries:
\begin{displaymath}
  \EMPTYSTRING \LEXICON [q] \SLASHF [t]
  \qquad
  \EMPTYSTRING \LEXICON [t] \SLASHF [a] \SLASHF [q']
\end{displaymath}
We also reuse the lexicon entries $\EMPTYSTRING \LEXICON [a]$ that we
introduced in Section~\ref{fragment:InitialConfiguration}.
Finally, we introduce the following rules:
\begin{align}
  \RULE{[q] \STACK \SLASHF [t]}{[t] \SLASHF X_1 \cdots \SLASHF X_m&}{[q] \STACK \SLASHF X_1 \cdots \SLASHF X_m}\label{rule:Existential1}
  \\
  \RULE{[t] \STACK \SLASHF [a']}{[a']&}{[t] \STACK}\label{rule:Existential2}
  \\
  \RULE{[t] \STACK \SLASHF [q']}{[q'] \SLASHF X_1 \cdots \SLASHF X_m&}{[t] \STACK \SLASHF X_1 \cdots \SLASHF X_m}\label{rule:Existential3}
\end{align}
A rule of the form~\eqref{rule:Existential1} is a composition rule of
degree~$m$ that simultaneously restricts the target of its primary
input to~$q$ and the target of its secondary input to~$t$.
A rule of the form~\eqref{rule:Existential2} is an application rule
that matches $t$ (the target of its primary input) with the tape
symbol $a'$ produced by~$t$ (its secondary input).
A rule of the form~\eqref{rule:Existential3} is a composition rule of
degree~$m$ that matches $t$ (the target of its primary input) with the
state $q'$ resulting from the application of~$t$.

\paragraph{Computational Complexity}

Again, each of the above lexical entries and rules has size in
$\order{m}$.
The number of rules of the form~\eqref{rule:Existential3} added to the
grammar is bounded by the possible choices of the transition $t$ ($q'$
is unique, given $t$), and is thus a polynomial function of
$\size{M}$.
Similar analyses apply to the other rules and lexical entries.  
We thus conclude that the overall contribution to $\size{G}$ in this
step is polynomial in the size of the input, and the construction can
be carried out in polynomial time, too.

\begin{figure}
  \small
  \vskip-\abovedisplayskip
  \begin{displaymath}
    \INFER{E(c) = \HIGHLIGHT{\CONTROL{q} \SLASHF [a] \SLASHF [b]}}{%
      \PROJECT[10]{[q] \SLASHF \PIMINUS}{\EMPTYSTRING}
      &
      \RDOTINFER[duplicate the tape stack]{\HIGHLIGHT{\PIMINUS \SLASHF [a] \SLASHF [b]}}{%
        \INFER{\HIGHLIGHT{\PIPLUS \SLASHF [a] \SLASHF [b] \SLASHF [a] \SLASHF [b]}}{%
          \INFER{\HIGHLIGHT{\PIPLUS \SLASHF [a] \SLASHF [b] \SLASHF \CONTROL{t_2}}}{%
            \INFER{\HIGHLIGHT{\PIPLUS \SLASHF [a] \SLASHF [b]}}{%
              \PROJECT[4]{\PIPLUS \SLASHF \CONTROL{t_1}}{\EMPTYSTRING}
              &
              \RDOTINFER[simulate $t_1$]{%
                \HIGHLIGHT{\CONTROL{t_1} \SLASHF [a] \SLASHF [b]}}{\SUBTREE{E(t_1(c)) = \HIGHLIGHT{\CONTROL{q_1} \SLASHF [b] \SLASHF [a_1]}}}%
            }
            &
            \PROJECT[5]{[b] \SLASHF [b] \SLASHF \CONTROL{t_2}}{\EMPTYSTRING}
          }
          &
          \RDOTINFER[simulate $t_2$]{\HIGHLIGHT{\CONTROL{t_2} \SLASHF [a] \SLASHF [b]}}{\SUBTREE{E(t_2(c)) = \HIGHLIGHT{\CONTROL{q_2} \SLASHF [b] \SLASHF [a_2]}}}%
        }
      }
    }
  \end{displaymath}
  \caption{Bird's"-eye view of the derivation fragment for a pair of
    transitions $\pi = (t_1, t_2)$ out of a universal configuration
    $(q, a b)$, where $t_1 = (q, a) \to (q_1, a_1)$ and
    $t_2 = (q, a) \to (q_2, a_2)$.}
  \label{fig:UniversalState}
\end{figure}

\subsubsection{Transitions Out of Universal Configurations}
\label{fragment:UniversalConfigurations}

We finally present the lexicon entries and rules needed to simulate
transitions out of universal configurations.
This is the most involved part of our construction.
To give an intuition, Figure~\ref{fig:UniversalState} provides a
bird's eye view of the fragment that our grammar derives for a
universal configuration of the form $c = (q, a b)$ and a pair of
transitions $\pi = (t_1, t_2)$ where $t_1 = (q, a) \to (q_1, a_1)$ and
$t_2 = (q, a) \to (q_2, a_2)$.
Recall that we may assume that every universal configuration has
exactly two applicable transitions.
On a high level, we construct two identical copies of the tape stack
and then simulate the transition $t_1$ on one copy and the transition
$t_2$ on the second copy.
This is done in such a way that both $t_1$ and $t_2$ must lead to
accepting configurations in order to terminate the derivation process.
The actual derivation in the fragment proceeds in three phases as
follows.
First, it duplicates the tape stack of the root category $E(c)$.
Second, it splits the duplicated stack into two identical halves, each
targeted at one of the two transitions.
Third, it simulates (in two separate branches) the two transitions on
their respective halves to arrive at the two leaf categories
$E(t_1(c))$ and $E(t_2(c))$.
(Note that the fragment here differs from the one in
Figure~\ref{fig:ExistentialState} in that it has \emph{two}
distinguished leaves, not one.)
In the following we describe the three phases in detail.

\paragraph{Phase~1: Duplicating the Tape Stack}

We illustrate this phase of the derivation in
Figure~\ref{fig:UniversalPhase1}.
The derivation starts by checking whether the root category $E(c)$
indeed encodes a universal configuration, and records this fact by
changing the target to the atomic category $\PIMINUS$ \eqref{rule:UniversalPartOne1}.
The intended interpretation of this category is that the derivation is
simulating the transition pair $\pi$ but has not yet duplicated the
tape stack (${-}$).
The derivation then nondeterministically extends the tape stack
\eqref{rule:UniversalPartOne2}, in much the same way as for the
initial configuration in Section~\ref{fragment:InitialConfiguration}.
At the end of the phase, the derivation tests whether the two halves
of the stack are equal, that is, whether the nondeterministic
extension indeed created an exact copy of the initial stack.
This test is crucial for our construction, and we describe it in more
detail below.
If the test is successful, the derivation changes the target to
$\PIPLUS$, signifying that the derivation has successfully duplicated
the tape stack.

To support derivations such as the one in
Figure~\ref{fig:UniversalPhase1}, we introduce the following lexicon
entries and rules.
Let $q \in Q$ be any universal state, and let $\pi = (t_1, t_2)$ be
any pair of transitions where $t_1 = (q, a) \to (q_1, a_1)$ and
$t_2 = (q, a) \to (q_2, a_2)$.
Let also $b \in \Sigma$ be any tape symbol.
We introduce the lexicon entry
$\EMPTYSTRING \LEXICON \CONTROL{q} \SLASHF \PIMINUS$ and reuse the
entries of the form $\EMPTYSTRING \LEXICON [b]$ that we introduced in
Section~\ref{fragment:InitialConfiguration}.
The following rules implement the change of the target of $E(c)$ and
nondeterministic extension of the tape stack:
\begin{align}
  \RULE{\CONTROL{q} \STACK \SLASHF \PIMINUS}{\PIMINUS \SLASHF X_1 \cdots \SLASHF X_m &}{\CONTROL{q} \STACK \SLASHF X_1 \cdots \SLASHF X_m}\label{rule:UniversalPartOne1}
  \\
  \RULE{\PIMINUS \STACK \SLASHF [b]}{[b] &}{\PIMINUS \STACK}\label{rule:UniversalPartOne2}
\end{align}
A rule of the form~\eqref{rule:UniversalPartOne1} is a composition
rule of degree~$m$ that simultaneously restricts the target of its
primary input to $\CONTROL{q}$ and the target of its secondary input
to $\PIMINUS$.
A rule of the form~\eqref{rule:UniversalPartOne2} is an application
rule that restricts the target of its primary input to $\PIMINUS$.

\begin{figure}
  \vskip-\abovedisplayskip
  \begin{displaymath}
    \INFER[\eqref{rule:UniversalPartOne1}]{\HIGHLIGHT{\CONTROL{q} \SLASHF [a] \SLASHF [b]}}{%
      \PROJECT[6]{\CONTROL{q} \SLASHF \PIMINUS}{\EMPTYSTRING}
      &
      \INFER[\eqref{rule:UniversalPartOne2}]{\HIGHLIGHT{\PIMINUS \SLASHF [a] \SLASHF [b]}}{%
        \INFER[\eqref{rule:UniversalPartOne2}]{\HIGHLIGHT{\PIMINUS \SLASHF [a] \SLASHF [b] \SLASHF [a]}}{%
          \RDOTINFER[equality test]{\HIGHLIGHT{\PIMINUS \SLASHF [a] \SLASHF [b] \SLASHF [a] \SLASHF [b]}}{\SUBTREE{\HIGHLIGHT{\PIPLUS \SLASHF [a] \SLASHF [b] \SLASHF [a] \SLASHF [b]}}}
          &
          \quad
          &
          \PROJECT[4]{[b]}{\EMPTYSTRING}
        }
        &
        \PROJECT[5]{[a]}{\EMPTYSTRING}
      }
    }
  \end{displaymath}
  \caption{Simulation of transitions out of universal configurations,
    phase~1: Duplicating the tape stack.}
  \label{fig:UniversalPhase1}
\end{figure}

It remains to describe how to implement the equality test.
To give an intuition, Figure~\ref{fig:EqualityTest} shows a derivation
that implements the test for the tape $ab$ of our running example.
For readability, we have numbered the arguments on the tape stack.
Step \eqref{rule:EqualityTest1} uses a composition rule of degree $2m$
to change the target of the root category to a new atomic category
$\PIEQ{1}$.
The intended interpretation of this category is that the derivation
needs to check whether the two halves of the tape stack agree at
position $1$ and $m+1$.
Accordingly, the composition used in step \eqref{rule:EqualityTest2}
is restricted in such a way that it can only be instantiated if the
two atomic categories at positions 1 and 3 are equal.
Similarly, the composition used in step \eqref{rule:EqualityTest3} can
only be instantiated if the two categories at positions 2 and 4 are
equal.
It is not hard to see that this can be scaled up to~$m$ tests, each of
which tests the equality of the categories at positions $i$ and
$m + i$.
Taken together, these tests check whether the two halves of the tape
are indeed identical.

More formally now, let $b \in \Sigma$ be any tape symbol.
We introduce the following shorthand notation, where $1 \leq i \leq m$
is any tape position:
\begin{displaymath}
  \eta_{i,b} \PAD{\equiv} \SLASHF X_1 \cdots \SLASHF X_{i-1} \SLASHF [b] \SLASHF X_{i+1} \cdots \SLASHF X_m \SLASHF X_{m+1} \cdots \SLASHF X_{m+i-1} \SLASHF [b] \SLASHF X_{m+i+1} \cdots \SLASHF X_{2m}
\end{displaymath}
Thus $\eta_{i,b}$ is a sequence of $2m$ slash"=variable pairs, except
that $X_i$ and $X_{m+i}$ have been replaced with the concrete atomic
category $[b]$.
Then to support derivations such as the one in
Figure~\ref{fig:EqualityTest}, we introduce the following lexicon
entries, where $1 \leq i < m$:
\begin{displaymath}
  \EMPTYSTRING \LEXICON \PIMINUS \SLASHF \PIEQ{1}
  \qquad
  \EMPTYSTRING \LEXICON \PIEQ{i} \SLASHF \PIEQ{i+1}
  \qquad
  \EMPTYSTRING \LEXICON \PIEQ{m} \SLASHF \PIPLUS
\end{displaymath}
We also introduce the following composition rules of degree~$2m$, for
$1 \leq i < m$:
\begin{align}
  \RULE{\PIMINUS \STACK \SLASHF \PIEQ{1}}{\PIEQ{1} \SLASHF X_1 \cdots \SLASHF X_{2m} &}{\PIMINUS \STACK \SLASHF X_1 \cdots \SLASHF X_{2m}}
  \label{rule:EqualityTest1}
  \\
  \RULE{\PIEQ{i} \STACK \SLASHF \PIEQ{i+1}}{\PIEQ{i+1} \eta_{i,b} &}{\PIEQ{i} \STACK \eta_{i,b}}
  \label{rule:EqualityTest2}
  \\
  \RULE{\PIEQ{m} \STACK \SLASHF \PIPLUS}{\PIPLUS \eta_{m,b} &}{\PIEQ{m} \STACK \eta_{m,b}}
  \label{rule:EqualityTest3}
\end{align}

\begin{figure}
  \vskip-\abovedisplayskip
  \begin{displaymath}
    \INFER[\eqref{rule:EqualityTest1}]{\HIGHLIGHT{\PIMINUS \SLASHF [a]_1 \SLASHF [b]_2 \SLASHF [a]_3 \SLASHF [b]_4}}{
      \PROJECT[3]{\PIMINUS \SLASHF \PIEQ{1}}{\EMPTYSTRING}
      &
      \INFER[\eqref{rule:EqualityTest2}]{\HIGHLIGHT{\PIEQ{1} \SLASHF [a]_1 \SLASHF [b]_2 \SLASHF [a]_3 \SLASHF [b]_4}}{%
        \PROJECT[2]{\PIEQ{1} \SLASHF \PIEQ{2}}{\EMPTYSTRING}
        &
        \INFER[\eqref{rule:EqualityTest3}]{\HIGHLIGHT{\PIEQ{2} \SLASHF [a]_1 \SLASHF [b]_2 \SLASHF [a]_3 \SLASHF [b]_4}}{%
          \PROJECT{\PIEQ{2} \SLASHF \PIPLUS}{\EMPTYSTRING}
          &
          \SUBTREE{\HIGHLIGHT{\PIPLUS \SLASHF [a]_1 \SLASHF [b]_2 \SLASHF [a]_3 \SLASHF [b]_4}}
        }
      }
    }
  \end{displaymath}
  \caption{Equality test for the tape $a b a b$.}
  \label{fig:EqualityTest}
\end{figure}

\paragraph{Phase~2: Splitting the Tape Stack}

In the second phase, the derivation branches off into two subtrees, as
was illustrated in Figure~\ref{fig:UniversalState}.
We present this second phase in more detail in
Figure~\ref{fig:UniversalStatePhase2}.
This derivation simulates the ``splitting'' of the tape stack into two
(identical) halves.
To implement it, we introduce lexicon entries
$\EMPTYSTRING \LEXICON \PIPLUS \SLASHF \CONTROL{t_1}$ and
$\EMPTYSTRING \LEXICON [b] \SLASHF [b] \SLASHF \CONTROL{t_2}$, where
$b \in \Sigma$ is any tape symbol.
We also introduce the following rules:
\begin{align}
  \RULE{\PIPLUS \STACK \SLASHF \CONTROL{t_2}}{\CONTROL{t_2} \SLASHF X_1 \cdots \SLASHF X_m &}{\PIPLUS \STACK \SLASHF X_1 \cdots \SLASHF X_m}\label{rule:Branching1}
  \\
  \RULE{\PIPLUS \STACK \SLASHF [b]}{[b] \SLASHF [b] \SLASHF \CONTROL{t_2} &}{\PIPLUS \STACK \SLASHF [b] \SLASHF \CONTROL{t_2}}\label{rule:Branching2}
  \\
  \RULE{\PIPLUS \STACK \SLASHF \CONTROL{t_1}}{\CONTROL{t_1} \SLASHF X_1 \cdots \SLASHF X_m &}{\PIPLUS \STACK \SLASHF X_1 \cdots \SLASHF X_m} \label{rule:Branching3}
\end{align}
Rule of the forms~\eqref{rule:Branching1} and~\eqref{rule:Branching3}
are composition rules of degree~$m$.
Note that this ensures that the categories targeted at $\CONTROL{t_1}$
and $\CONTROL{t_2}$ encode a tape of~$m$ symbols.
A rule of the form~\eqref{rule:Branching2} is a composition rule of
degree~$2$.

\begin{figure}
  \vskip-\abovedisplayskip
  \begin{displaymath}
    \INFER[\eqref{rule:Branching1}]{\HIGHLIGHT{\PIPLUS \SLASHF [a] \SLASHF [b] \SLASHF [a] \SLASHF [b]}}{%
      \INFER[\eqref{rule:Branching2}]{\HIGHLIGHT{\PIPLUS \SLASHF [a] \SLASHF[b] \SLASHF \CONTROL{t_2}}}{%
        \INFER[\eqref{rule:Branching3}]{\HIGHLIGHT{\PIPLUS \SLASHF [a] \SLASHF [b]}}{%
          \PROJECT{\PIPLUS \SLASHF \CONTROL{t_1}}{\EMPTYSTRING}
          &
          \SUBTREE{\HIGHLIGHT{\CONTROL{t_1} \SLASHF [a] \SLASHF [b]}}
        }
        &
        \PROJECT[2]{[b] \SLASHF [b] \SLASHF \CONTROL{t_2}}{\EMPTYSTRING}
      }
      &
      \SUBTREE{\HIGHLIGHT{\CONTROL{t_2} \SLASHF [a] \SLASHF [b]}}
    }
  \end{displaymath}
  \caption{Simulation of transitions out of universal configurations,
    phase~2: Splitting the tape stack.}
  \label{fig:UniversalStatePhase2}
\end{figure}

\paragraph{Phase~3: Simulating the Two Transitions}

In the third and final phase, the derivation simulates the two
transitions $t_1$ and $t_2$.
To implement this phase we do not need to introduce any new lexicon
entries or rules; we can simply reuse part of the construction that we
presented in Section~\ref{fragment:ExistentialConfigurations} for the
existential states.
More specifically, we can reuse the part of that construction that
starts with a category with target $\CONTROL{t}$ and uses
rules~(\ref{rule:Existential2}) and~(\ref{rule:Existential3}).

\paragraph{Computational Complexity}

All of the introduced lexical entries have size bounded by some
constant independent of the input size.
At the same time, all of the rules introduced in the four phases above
have degree bounded by~$2m$.
It is easy to see that we can construct each of these elements in time
$\order{m}$.
Furthermore, the number of lexical entries and rules produced in this
step is bounded by a polynomial function of the input size.
For instance, we add to $G$ a number $\size{\Sigma} \cdot m$ of rules
of types~\eqref{rule:EqualityTest1} and~\eqref{rule:EqualityTest2},
since there is a single rule for each choice of a tape symbol $a$ and
index $i$ with $1 \leq i < m$.
Similar analyses can be carried out for the remaining elements.
We then conclude that the overall contribution to $\size{G}$ in this
step is polynomial in $\size{M}$ and $\size{w}$, and the construction
can be carried out in the same amount of time.

\subsubsection{Correctness}

With all of the grammar components in place, we now address the
correctness of our construction.
We argue that the sentential derivations of~$G$ exactly correspond to
the accepting computations of~$M$ when applied to the input
string~$w$.
To do this, we read $G$'s derivations in the canonical direction, that
is, from the leaves to the root.
First of all, observe that the fragments introduced in the various
steps of the construction all use reserved target categories, and they
all use rules with target restrictions for these categories.
In this way, it is not possible in a derivation to mix fragments from
different steps---that is, fragments cannot be broken apart.

\paragraph{Accepting Configurations}

A (sentential) derivation in $G$ starts with the fragments introduced
in Section~\ref{fragment:AcceptingConfigurations}.
Each of these fragments uses composition rules to combine several tape
symbols into a category of the form $[\CMD{accept}] \alpha$, and then
switches to a category of the form $[q] \alpha$ with $g(q) = \ACCEPT$.
Because of the degree restriction of Rule~\eqref{rule:Accept1}, the
switch is only possible if $\alpha$ has exactly $m$ arguments; in all
other cases the derivation will come to a dead end, that is, it will
not derive the distinguished category of the grammar.
The categories $[q] \alpha$ composed by the fragments of
Section~\ref{fragment:AcceptingConfigurations} encode accepting
configurations of~$\myTM$, and it is not difficult to see that all
possible accepting configurations with $g(q) = \ACCEPT$ can be
generated by these fragments.
These are exactly the leaves of our valid computation trees.

\paragraph{Existential Configurations}

The derivation freely attempts to apply the transitions of~$\myTM$ to
the categories obtained as above and, recursively, to all the
categories that result from the application of these transitions.
More precisely, a transition $t$ applying to an existential
configuration is simulated (in reverse) on a category $[q] \alpha$
using the fragment of
Section~\ref{fragment:ExistentialConfigurations}.
This is done using Rule~\eqref{rule:Existential3}, which switches from
a category with target $[q]$ to a category with target $[t]$, and
produces a new category whose stack has $m+1$ arguments.
At this point, only some rule of type~\eqref{rule:Existential2} can be
applied, resulting in the reduction of the stack, immediately followed
by some rule of type~\eqref{rule:Existential1}, which is a composition
rule of degree~$m$.
If the derivation were to use more than one occurrence
of~\eqref{rule:Existential2}, then it would derive a category whose
stack contains fewer than $m$ elements.
As a consequence, Rule~\eqref{rule:Existential1} would no longer be
applicable, because of the restriction on the composition degree, and
the whole derivation would come to a dead end.

\paragraph{Universal Configurations}

The derivation can also simulate (in reverse) the two transitions
$t_1$ and $t_2$ applying to a universal state $q$.
This is done using the fragments of
Section~\ref{fragment:UniversalConfigurations}.
In this case the derivation starts with
Rules~\eqref{rule:Existential3} and~\eqref{rule:Existential2} used for
the existential states; but now the involved categories have targets
$[t_i]$ disjoint from the targets used for the existential states,
since transitions $t_i$ apply to universal states.
The simulation of $t_1$ and $t_2$ results in categories $[q] \alpha$
and $[q] \alpha'$ with the same target $[q]$ and with $m$ arguments
each.
These categories are then merged into a new category
$[q] \alpha \alpha'$ by concatenating their stacks, and an equality
test is successively carried out on $\alpha \alpha'$.
If the test is successful, the derivation pops an arbitrary number of
arguments from $[q] \alpha \alpha'$, resulting in a new category of
the form $[q] \alpha''$.
Rule~\eqref{rule:UniversalPartOne1} can then be applied only in case
$[q] \alpha''$ has exactly $m$ arguments.
This means that $[q] \alpha''$ encodes one of the configurations
of~$\myTM$, and that $\alpha = \alpha' = \alpha''$.

\paragraph{Initial Configuration}

Finally, if the above process ever composes a category $[q_0] \alpha$
encoding the initial configuration of~$\myTM$ relative to the input
string $w$, then the derivation uses the fragment of
Section~\ref{fragment:InitialConfiguration}.
The rules of this fragment switch the target category from $[q_0]$ to
$\CONTROL{\CMD{init}}$, the distinguished category of~$G$, and then
pop the stack arguments, providing thus a sentential derivation
for~$\EMPTYSTRING$.

\bigskip

This correctness argument finally concludes the proof of our
Lemma~\ref{lem:EXPTIMEHardness}.

\subsection{Membership in \EXPTIME}
\label{sec:EXPTIMEMembership}

It remains to prove the following:

\begin{lemma}
  The universal recognition for unrestricted \VWCCG\ is in \EXPTIME.
\end{lemma}

\noindent To show this, we extend an existing recognition algorithm by
\citet{kuhlmann2014new} that takes as input a \VWCCG\ $G$ with no
empty categories and a string $w$, and decides whether $w \in L(G)$.

\paragraph{Complexity of the Algorithm of Kuhlmann and Satta}

The algorithm of \citet{kuhlmann2014new} is based on a special
decomposition of \CCG\ derivations into elementary pieces, adapting an
idea first presented by \citet{vijayshanker1990polynomial}.
These elementary pieces are specially designed to satisfy two useful
properties: First, each elementary piece can be stored using an amount
of space that does not depend on the length of~$w$.
Second, elementary pieces can be shared among different derivations
of~$w$ under~$G$.
The algorithm then uses dynamic programming to construct and store in
a multi"-dimensional parsing table all possible elementary pieces
pertaining to the derivations of~$w$ under~$G$.
From such table one can directly check whether $w \in L(G)$.
Despite the fact that the number of derivations of~$w$ under~$G$ can
grow exponentially with the length of~$w$, the two properties of
elementary pieces allow the algorithm to run in time polynomial in the
length of~$w$.
However, the runtime is not bounded by a polynomial function in the
size of~$G$, as should be expected from the hardness results reported
in Section~\ref{sec:NPHardness}.

More specifically, let $\ATOMIC$ be the set of all atomic categories
of the input grammar $G$, and let $\LEXARGS$ be the set of all
arguments occurring in the categories in $G$'s lexicon.
Let also~$d$ be the maximum degree of a composition rule in~$G$,
let~$a$ be the maximum arity of an argument in $\LEXARGS$, and
let~$\ell$ be the maximum number of arguments in the categories in
$G$'s lexicon.
We set $c_G = \max \SET{d+a, \ell}$.
\citet{kuhlmann2014new} report for their algorithm a running time in
$\order{\size{\ATOMIC} \cdot \size{\LEXARGS}^{2 c_G} \cdot
  \size{w}^6}$.

To see that this upper bound is an exponential function in the size of
the input, observe that the quantities $\size{\ATOMIC}$ and
$\size{\LEXARGS}$ are both bounded by $\size{G}$, since each category
in $\ATOMIC$ or in $\LEXARGS$ must also occur in~$G$.
Furthermore, $\ell$ is bounded by the maximum length of a category in
$G$'s lexicon, and thus by $\size{G}$.
Similarly, $d$ is bounded by the length of some secondary component in
a composition rule of~$G$, and $a$ is bounded by the length of some
category in $G$'s lexicon.
Then $d+a$ is bounded by $\size{G}$ as well.
Combining the previous observations, we have that $c_G$ is bounded
by~$\size{G}$.
We can then conclude that the runtime of the recognition algorithm is
bounded by
\begin{equation}
  \label{eq:KuhlmannSattaComplexity}
  \size{G} \cdot \size{G}^{2|G|} \cdot \size{w}^6 \PAD[\;]{=}
  \size{G}^{1+2|G|} \cdot \size{w}^6 \PAD[\;]{=}
  2^{\log{\size{G}} + 2\size{G}\log{\size{G}}} \cdot \size{w}^6\,,
\end{equation}
which is indeed exponential in the size of the input.  

\paragraph{Extension of the Algorithm of Kuhlmann and Satta to \VWCCG}

As already mentioned, the algorithm by \citet{kuhlmann2014new} works
for a grammar $G$ with no empty categories.
More specifically, the algorithm starts by adding to the parsing table
items of the form $[X, i, i + 1]$ for each category $X$ that is
assigned by $G$'s lexicon to the $i$-th word in the input string $w$.
Here $[X, i, i + 1]$ represents an elementary piece of derivation
consisting of a tree with two nodes: a root with label $X$ and a child
node with label the $i$-th word of~$w$.
In order to extend the algorithm to unrestricted \VWCCG, all we need
to do is add to the parsing table items of the form $[X, i, i]$ for
every empty category $X$ in $G$'s lexicon and for every integer $i$
with $0 \leq i \leq \size{w}$.
This creates new elementary pieces of derivations accounting for the
empty string.
These pieces can be combined with each other, as well as with other
pieces already existing in the table, triggering the construction of
derivations that involve empty categories.

The proof of the correctness of the recognition algorithm by
\citet{kuhlmann2014new} immediately extends to the new algorithm.
This is so because the proof only rests on structural properties of
\CCG\ derivations, without any assumption about the fact that these
derivations involve words from~$w$ or the empty string.
Furthermore, the exponential runtime reported above still holds for
the new algorithm.
This is a consequence of the fact that we use the same item
representation as in the original algorithm for the elementary pieces
of derivations involving the empty string.

While the algorithms discussed above are designed for the random
access machine architecture, or RAM for short, it is well"-known that
any algorithm working on a RAM can be computed on a deterministic
Turing machine with only polynomial"-time overhead; see for instance
\citet[Theorem~2.5]{papadimitriou1994computational}.
We can thus conclude that the universal recognition problem for
\VWCCG\ can still be solved in exponential time on a deterministic
Turing machine.

\subsection{Discussion}
\label{sec:EXPTIMEDiscussion}

In view of our proof of Theorem~\ref{thm:EXPTIMECompleteness}, we now
come back to the question raised in Section~\ref{sec:NPDiscussion}.
Specifically, we want to further investigate the features that are
responsible for the additional complexity that comes with unrestricted
\VWCCG.
In our reduction in Section~\ref{sec:EXPTIMEHardness} we have used two
features of the formalism that were already discussed in
Section~\ref{sec:NPDiscussion}, namely the capability to define rules
with restrictions on their secondary input categories, and the
capability to define rules whose secondary input categories do not
have any constant bound on their arity.
In addition, we have also exploited two new features, listed below.
Again, dropping any one of these four features would break our proof
(but see our discussion in Section~\ref{sec:UnboundedComposition}).

\paragraph{Derivational Ambiguity}

Our grammar makes crucial use of derivational ambiguity.
More precisely, $G$ encodes $\myTM$'s configurations relative to $w$
into its own categories.
We assign all of these categories to the empty string $\ep$, thus
introducing massive ambiguity in $G$'s derivations.
Our proof of Lemma~\ref{lem:EXPTIMEHardness} would not work if we
restricted ourself to the use of unambiguous grammars, and we note
that the computational complexity of the universal recognition problem
for the class \VWCCG\ restricted to unambiguous grammars is currently
unknown.
Furthermore, on a par with lexical ambiguity discussed in
Section~\ref{sec:NPDiscussion}, syntactic ambiguity is an essential
feature in most formalisms for the modeling of natural language
syntax, including those whose universal recognition problem can be
parsed in polynomial time, such as \TAG{}s.
As such, this individual feature cannot be held responsible, at least
in isolation, for the complexity of the recognition problem for the
class \VWCCG, and in designing a polynomially parsable version of
\VWCCG\ we would not be interested in blocking derivational ambiguity.

\paragraph{Unlexicalized Rules}

We remind the reader that, broadly speaking, a lexicalized rule in a
string rewriting formalism is a rule that (directly) produces some
lexical token.
The rule is also thought to be specialized for that token, meaning that
the rule contributes to the derivation by introducing some structure
representing the syntactic frame and valencies of the token itself.
The basic idea of our proof is to simulate valid computations
of~$\myTM$ on~$w$ through derivations of~$G$.
In particular, each fragment in a derivation of $G$ uniquely
represents some node of a valid computation.
It is therefore essential in our proof that each fragment uses
unlexicalized rules, that is, generates the empty string.
One may view this phenomenon in the light of computational complexity.
It is not difficult to verify that given an unlexicalized grammar $G$,
it cannot be transformed into a lexicalized grammar $G'$ in polynomial
time unless NP $=$ EXPTIME.\footnote{This equality is considered
  unlikely to hold since it would, for instance, imply that NP $=$
  PSPACE, and that the polynomial hierarchy collapses.}
For assume that such a polynomial"-time transformation $T$ exists. 
This would imply that an arbitrary instance $(G,w)$ of the universal
recognition problem with $\EMPTYSTRING$-entries can be converted (in
polynomial time) into an equivalent instance $(T(G),w)$ without
$\EMPTYSTRING$-entries.
We know that the former problem is EXPTIME"-complete and the latter
problem is NP"-complete.
Thus, the existence of $T$ imply that NP $=$ EXPTIME.
One should observe that this is not an effect of $T(G)$ being
prohibitively large: the size of the lexicalized grammar $T(G)$ is
polynomially bounded in the size of $G$, since $T$ is computable in
polynomial time.

\section{General Discussion}
\label{sec:Discussion}

The computational effect of grammar structure and grammar size on the
parsing problem is rather well understood for several formalisms
currently used in computational linguistics, including context"-free
grammar and \TAG.
However, to the best of our knowledge, this problem has not been
investigated before for \VWCCG\ or other versions of \CCG; see for
instance \citet{kuhlmann2014new} for discussion.
In this article we have shed some light on the impact of certain
features of \VWCCG\ on the computational complexity of the parsing
problem.
We have shown that the universal recognition problem for \VWCCG\ is
dramatically more complex than the corresponding problem for \TAG,
despite the already mentioned weak equivalence between these two
formalisms.
Our results therefore solve an open problem for \VWCCG\ and raise
important questions about the computational complexity of contemporary
incarnations of \CCG.
In this section we would like to conclude the article with a
discussion of our results in the broader context of current research.

\subsection{Sources of Complexity}

The two features of \VWCCG\ that are at the core of our complexity
results are the rule restrictions and the unbounded degree of
composition rules.
As already mentioned, dropping any one of these two features would
break our specific constructions.
At the same time, it is important to consider the problem from the
dual perspective: we do not know whether dropping \emph{any
  combination} of these two features would admit a polynomial"-time
parsing algorithm for \VWCCG---and this holds true regardless of the
grammars being $\EMPTYSTRING$"-free or not.
Perhaps most important for current practice, this means that we do not
know whether modern versions of \CCG\ can be parsed in polynomial
time.
To illustrate the point, the algorithm by \citet{kuhlmann2014new}
(Section~\ref{sec:EXPTIMEMembership}) takes as input an
$\EMPTYSTRING$"-free \VWCCG\ $G$ and a string $w$, and decides whether
$w \in L(G)$.
Even if $G$ has no rule restrictions and the degree of its composition
rules is considered as a constant, the upper bound that we would get
from the analysis in Equation~\eqref{eq:KuhlmannSattaComplexity} would
still be exponential in the grammar size.
This shows that our understanding of the computational properties of
\CCG\ is still quite limited.

\paragraph{Epsilon Entries}

One important issue that needs further discussion here is the role of
$\EMPTYSTRING$"-entries in \VWCCG.
From the linguistic perspective, $\EMPTYSTRING$"-entries violate one
of the central principles of \CCG, the Principle of Adjacency
\citep[p.~54]{steedman2000syntactic}.
From the computational perspective, $\EMPTYSTRING$"-entries represent
the boundary between the results in Section~\ref{sec:NPCompleteness}
and Section~\ref{sec:EXPTIMECompleteness}.
However, since we do not know whether the classes NP and EXPTIME can
be separated, we cannot draw any precise conclusion about the role of
$\EMPTYSTRING$"-entries in the parsing problem.
Even under the generative perspective, we do not know the exact role
of $\EMPTYSTRING$"-entries.
More precisely, the proof of the weak equivalence between \VWCCG\ and
\TAG\ provided by \citet{vijayshanker1994equivalence} makes crucial
use of $\EMPTYSTRING$"-entries, and it is currently unknown whether
the generative capacity of \VWCCG\ without $\EMPTYSTRING$"-entries is
still the same as that of \TAG, or if it is strictly smaller.
This is another important open problem that attests our lack of
theoretical understanding of \CCG.

\paragraph{Unbounded Composition}\label{sec:UnboundedComposition}

A second issue that we would like to discuss is related to the notion
of degree of composition rules in \VWCCG.
According to the original definition of \VWCCG, each \emph{individual}
grammar in this class has a specific bound on the degree of its
composition rules, but there is no constant bound holding for all
grammars.
As already discussed in Sections~\ref{sec:NPDiscussion}
and~\ref{sec:EXPTIMEDiscussion}, the complexity results in this
article do exploit this property in a crucial way.
However, there are two alternative scenarios that we want to consider
here.
In a first scenario, one could state that there exists some
language"-independent constant that bounds the degree of composition
rules \emph{for all} grammars in the class \VWCCG.
This would break all of the constructions in this article.
The second possible scenario is one that has been discussed by, among
others, \citet[Section~5.2]{weir1988combinatory} and
\citet[p.~210]{steedman2000syntactic}: We could define a formalism
alternative to \VWCCG, in which an individual grammar is allowed to
use composition rules of unbounded degree.
This would mean that the $\STACK$ notation introduced in
Equation~(\ref{rule:RestrictedBackwardCrossedComposition2}) must be
used in the primary category as well as in the secondary category of a
composition rule.
Such a move would go into the opposite direction with respect to the
first scenario above, reaching the power of Turing machines, as
informally explained in what follows.
Recall that in Section~\ref{sec:EXPTIMECompleteness} we have used
\VWCCG\ derivations to simulate moves of an ATM working with a
circular tape whose size is bounded by some polynomial function of the
length of the input. 
Specifically, we have encoded such a tape into some category $X$, and
have used $X$ as a primary or as a secondary input category in
composition rules, in order to simulate the moves of the ATM.
If we now allow the use of composition rules of arbitrarily large
degree within an individual grammar, we can simulate the moves of a
general Turing machine, in a way very similar to what we have done
with our ATMs.
This shows that the degree of composition rules can play a very
powerful role in the definition of \CCG\ formalisms.

\paragraph{A Note on Worst"-Case Analysis}

Our analysis of parsing complexity examines how the parser will
perform in the least favourable situations.
This perspective is justified by our interest in the question of where
\CCG\ sits within the landscape of mildly context"-sensitive grammars,
which are characterized by worst"-case polynomial"-time parsing
\citep{joshi1985tree}.
On the other hand, our results do not allow us to draw strong
conclusions about practical average"-case or expected parsing
complexity, a question that many practitioners in the field may be
more interested in when choosing a formalism for a problem.
At the same time, recent progress on the development of practical
\CCG\ parsers has shown that with suitable heuristics, this formalism
can be processed with very high efficiency
\citep{Lewis14a*ccg,Lee-emnlp-2016}.
We tend to view empirical and worst"-case complexity as two orthogonal
issues, where the latter enriches our understanding of the problem and
might lead to the development of new, improved formalisms and
algorithms, often with further advancements on the practical side.

\subsection{Succinctness}

As already mentioned \VWCCG\ is known to be generatively equivalent to
\TAG, in the weak sense, as shown by \citet{weir1988combinatory} and
\citet{vijayshanker1994equivalence}.  
\citet{schabes1990mathematical} reports that the universal recognition
problem for \TAG\ can be decided in time
$\order{\size{G}^2 \size{w}^6}$, where $\size{G}$ is the size of the
input grammar $G$ and $\size{w}$ is the length of the input sentence
$w$.
One could hope then to efficiently solve the universal recognition
problem for \VWCCG\ by translating an input \VWCCG\ $G$ into an
equivalent \TAG\ $G'$, and then applying to $G'$ and the input string
any standard recognition method for the latter class.
However, the part of the equivalence proof by
\citet{vijayshanker1994equivalence} showing how to translate \VWCCG\
to \TAG\ requires the instantiation of a number of elementary trees in
$G'$ that is exponential in $\size{G}$.
(Trees are the elementary objects encoding the rules in a \TAG.)

The fact that the same class of languages can be generated by grammar
formalisms with substantially different parsing complexity naturally
leads us to the notion of the \emph{succinctness} of a grammar,
cf.~\citet{Hartmanis:sicomp80}.
In formal language theory, grammar succinctness is used to measure the
expressive capacity of a grammar formalism, as opposed to its
generative capacity.
More precisely, grammar succinctness measures the amount of resources
that different grammar formalisms put in place in order to generate
the same language class.
As a simple example, it is well known that certain finite languages
can be generated by context"-free grammars that are very compact, that
is, small in size, while the same languages require finite state
automata of size exponentially larger.
In computational linguistics, succinctness was first discussed in the
context of the formalism of ID/LP-grammar, a variant of context"-free
grammar where the ordering of the nonterminals in the right-hand side
of a rule can be relaxed.
\citet{moshier1987succinctness} show that ID/LP-grammars are
exponentially more succinct than context"-free grammars.
As in the example above, this means that there are languages for which
any context"-free grammar must necessarily be at least
super"-polynomially larger than the smallest ID/LP-grammar.
A similar fact holds for \VWCCG: By our result in
Section~\ref{sec:NPCompleteness} there are languages for which there
exist small \VWCCG{s} but where the weakly equivalent \TAG\ must
necessarily be at least exponentially larger (unless PTIME~$=$~NP).
If we allow $\EMPTYSTRING$-entries, then
Section~\ref{sec:EXPTIMECompleteness} provides a stronger result: we
can get rid of the qualification ``unless PTIME~$=$~NP'', as PTIME
$\neq$ EXPTIME holds unconditionally (cf.\ \citealp[Theorem
7.1]{papadimitriou1994computational} and the subsequent corollary).
Since we can also translate any \TAG\ into an equivalent \VWCCG\
without blowing up the size of the grammar, following the construction
by \citet{vijayshanker1994equivalence}, we conclude that \VWCCG\ is
more succinct than \TAG.
However, the price we have to pay for this gain in expressivity is the
extra parsing complexity of \VWCCG.

\subsection{The Landscape of Mildly Context-Sensitive Grammars}

Finally, connecting back to the original motivation of this work that we gave in Section~\ref{sec:Introduction}, we would like to conclude our discussion by placing our
results for \VWCCG\ in the broader scenario of the class of mildly
context"-sensitive formalisms.
This provides a more complete picture of this class than what we had
before.
The (informal) class of mildly context"-sensitive grammar formalisms
had originally been proposed by \citet{joshi1985tree} to provide
adequate descriptions of the syntactic structure of natural language.
This class includes formalisms whose generative power is only slightly
more powerful than context"-free grammars, that is, far below the one
of context"-sensitive grammars.

\begin{figure}
  \centering\small
  \newcommand*{\ENTRY}[3]{\node[fill=white] at (#1,#2) {\begin{tabular}{c}#3\end{tabular}};}%
  \begin{tikzpicture}[x=3cm,y=1.5cm]
    \draw[step=1,gray] (0, 0) grid (3, 3);
    \ENTRY{0}{0}{CFG}
    \ENTRY{1}{0}{TAG};
    \ENTRY{2}{0}{wn-LCFRS$(k)$};
    \ENTRY{0}{1}{\strut ID/LP \\ \strut grammar};
    \ENTRY{1}{1}{\strut $\EMPTYSTRING$"-free \\ \strut \VWCCG\ (?)};
    \ENTRY{3}{1}{LCFRS$(k)$};
    \ENTRY{3}{2}{LCFRS};
    \ENTRY{1}{3}{\VWCCG};
    \ENTRY{3}{3}{MCFG};
    \node[anchor=east] at (-0.5, 3) {\strut \textbf{EXPTIME}};
    \node[anchor=east] at (-0.5, 2) {\strut \textbf{PSPACE}};
    \node[anchor=east] at (-0.5, 1) {\strut \textbf{NP}};
    \node[anchor=east] at (-0.5, 0) {\strut \textbf{PTIME}};
    \node[anchor=north] at (0, -0.5) {\strut \textbf{CFL}};
    \node[anchor=north] at (1, -0.5) {\strut \textbf{TAL}};
    \node[anchor=north] at (2, -0.5) {\strut \textbf{wn-LCFRL}};
    \node[anchor=north] at (3, -0.5) {\strut \textbf{LCFRL}};
  \end{tikzpicture}
  \caption{Weak generative capacity (horizontal axis) versus 
  computational complexity (vertical axis) of various 
  mildly context"-sensitive grammar formalisms.}
  \label{fig:BigPicture}
\end{figure}

In Figure~\ref{fig:BigPicture} we map several known mildly
context"-sensitive formalisms into a two"-dimensional grid defined by
the generative capacity of the formalism (horizontal axis) and the
computational complexity of the universal recognition problem
(vertical axis).
For comparison, we also include in the picture some formalisms
generating the context"-free languages.
We thus start at the leftmost column of the grid with the class of
context"-free grammar and the class of ID/LP grammar.
As already mentioned, while these two classes are generatively
equivalent, ID/LP grammar is more succinct than context"-free grammar
and, as a consequence, the two classes do not have the same
computational complexity.
On the next column to the right, we reach the generative power of
tree"-adjoining languages, which is strictly larger than that of
context"-free languages.
Both \TAG\ and \VWCCG\ are in this column but, by the results in this
article, the computational complexity of \VWCCG\ is far above the one
of \TAG, again due to the increase in expressivity for the latter
class.
We have also tentatively placed $\EMPTYSTRING$"-free \VWCCG\ in this
column, although we do not know at this time whether the generative
capacity of this class is the same as that of general \VWCCG, hence
our question mark in the figure. 
In the next column to the right we find the class of well"-nested
linear context"-free rewriting system with fan-out bounded by~$k$,
written wn-LCFRS$(k)$.
A rewriting system in wn-LCFRS$(k)$ generates languages of string
tuples, where the number of components in each tuple is bounded by a
constant $k$ called the fan-out of the system.
The system exploits rules that work by combining tuple components in a
way that satisfies the so"-called \emph{well"-nestedness} condition, a
generalization of the standard condition on balanced brackets.
While this class further extends the generative capacity of \TAG\ (as
a special case, the class wn-LCFRS$(2)$ is generatively equivalent to
\TAG), it manages to keep the complexity of the universal recognition
problem in PTIME, as shown by \citet{gomez2010efficient}.
In the last column of our grid we have placed the class of linear
context"-free rewriting system (LCFRS) and the class of LCFRS with
fan-out bounded by a constant $k$ (LCFRS$(k)$), which have been
originally defined by \citet{vijayshanker1987characterizing}.
Historically, LCFRS has been introduced before wn-LCFRS$(k)$, and the
latter class was investigated as a restricted version of LCFRS.
In this column we also find the class of multiple context"-free
grammar (MCFG) defined by \citet{seki1991multiple}, who also prove the
generative equivalence result with LCFRS.
The computational complexity results displayed in this column are from
\citet{kaji1992universal} and \citet{satta1992recognition}.
All these systems generate string tuples but they do not satisfy the
well"-nestedness condition of wn-LCFRS$(k)$.
As a result, even in case of the class LCFRS$(k)$ where the fan-out is
bounded by a constant $k$, these systems cannot be parsed in
polynomial time, unless PTIME~$=$~NP, in contrast with the class
wn-LCFRS$(k)$.

\twocolumn

\bibliographystyle{compling}
\bibliography{mcqm2,mcqm}

\begin{thebibliography}{35}
\expandafter\ifx\csname natexlab\endcsname\relax\def\natexlab#1{#1}\fi

\bibitem[{Baldridge(2002)}]{baldridge2002lexically}
Baldridge, Jason. 2002.
\newblock \emph{Lexically Specified Derivational Control in Combinatory
  Categorial Grammar}.
\newblock Ph.D. thesis, University of Edinburgh, Edinburgh, UK.

\bibitem[{Baldridge and Kruijff(2003)}]{baldridge2003multimodal}
Baldridge, Jason and Geert-Jan~M. Kruijff. 2003.
\newblock {Multi}-{Modal} {Combinatory} {Categorial} {Grammar}.
\newblock In \emph{Tenth Conference of the European Chapter of the Association
  for Computational Linguistics (EACL)}, pages 211--218, Budapest, Hungary.

\bibitem[{Chandra, Kozen, and Stockmeyer(1981)}]{chandra1981alternation}
Chandra, Ashok~K., Dexter~C. Kozen, and Larry~J. Stockmeyer. 1981.
\newblock Alternation.
\newblock \emph{Journal of the Association for Computing Machinery},
  28(1):114--133.

\bibitem[{Clark and Curran(2007)}]{clark2007wide}
Clark, Stephen and James~R. Curran. 2007.
\newblock Wide-coverage efficient statistical parsing with {CCG} and log-linear
  models.
\newblock \emph{Computational Linguistics}, 33(4):493--552.

\bibitem[{Gazdar(1987)}]{gazdar1987applicability}
Gazdar, Gerald. 1987.
\newblock Applicability of indexed grammars to natural language.
\newblock In Uwe Reyle and Christian Rohrer, editors, \emph{Natural Language
  Parsing and Linguistic Theories}. D. Reidel, pages 69--94.

\bibitem[{G{\'o}mez-Rodr{\'\i}guez, Kuhlmann, and
  Satta(2010)}]{gomez2010efficient}
G{\'o}mez-Rodr{\'\i}guez, Carlos, Marco Kuhlmann, and Giorgio Satta. 2010.
\newblock Efficient parsing of well-nested linear context-free rewriting
  systems.
\newblock In \emph{Proceedings of Human Language Technologies: The 2010 Annual
  Conference of the North American Chapter of the Association for Computational
  Linguistics (NAACL)}, pages 276--284, Los Angeles, USA.

\bibitem[{Hartmanis(1980)}]{Hartmanis:sicomp80}
Hartmanis, Juris. 1980.
\newblock On the succinctness of different representations of languages.
\newblock \emph{{SIAM} J. Comput.}, 9(1):114--120.

\bibitem[{Hockenmaier and Steedman(2007)}]{hockenmaier2007ccgbank}
Hockenmaier, Julia and Mark Steedman. 2007.
\newblock {CCGbank}. {A} corpus of {CCG} derivations and dependency structures
  extracted from the {Penn Treebank}.
\newblock \emph{Computational Linguistics}, 33:355--396.

\bibitem[{Je{\.z} and Okhotin(2011)}]{jez2011complexity}
Je{\.z}, Arthur and Alexander Okhotin. 2011.
\newblock Complexity of equations over sets of natural numbers.
\newblock \emph{Theory of Computing Systems}, 48(2):319--342.

\bibitem[{Joshi(1985)}]{joshi1985tree}
Joshi, Aravind~K. 1985.
\newblock {Tree Adjoining Grammars}: How much context-sensitivity is required
  to provide reasonable structural descriptions?
\newblock In David~R. Dowty, Lauri Karttunen, and Arnold~M. Zwicky, editors,
  \emph{Natural Language Parsing}. Cambridge University Press, pages 206--250.

\bibitem[{Joshi and Schabes(1997)}]{joshi1997tree}
Joshi, Aravind~K. and Yves Schabes. 1997.
\newblock {Tree-Adjoining Grammars}.
\newblock In Grzegorz Rozenberg and Arto Salomaa, editors, \emph{Handbook of
  Formal Languages}, volume~3. Springer, pages 69--123.

\bibitem[{Kaji et~al.(1992)Kaji, Nakanishi, Seki, and
  Kasami}]{kaji1992universal}
Kaji, Yuichi, Ryuichi Nakanishi, Hiroyuki Seki, and Tadao Kasami. 1992.
\newblock The universal recognition problems for multiple context-free grammars
  and for linear context-free rewriting systems.
\newblock \emph{IEICE Transactions on Information and Systems},
  E75-D(1):78--88.

\bibitem[{Kuhlmann, Koller, and Satta(2015)}]{kuhlmann2015lexicalization}
Kuhlmann, Marco, Alexander Koller, and Giorgio Satta. 2015.
\newblock Lexicalization and generative power in {CCG}.
\newblock \emph{Computational Linguistics}, 41(2):187--219.

\bibitem[{Kuhlmann and Satta(2014)}]{kuhlmann2014new}
Kuhlmann, Marco and Giorgio Satta. 2014.
\newblock A new parsing algorithm for {Combinatory Categorial Grammar}.
\newblock \emph{Transactions of the Association for Computational Linguistics},
  2(Oct):405--418.

\bibitem[{Lee, Lewis, and Zettlemoyer(2016)}]{Lee-emnlp-2016}
Lee, Kenton, Mike Lewis, and Luke Zettlemoyer. 2016.
\newblock Global neural {CCG} parsing with optimality guarantees.
\newblock In \emph{Proceedings of the 2016 Conference on Empirical Methods in
  Natural Language Processing}, pages 2366--2376, Association for Computational
  Linguistics.

\bibitem[{Lewis and Steedman(2013)}]{Lewis-emnlp-2013}
Lewis, Mike and Mark Steedman. 2013.
\newblock Unsupervised induction of cross-lingual semantic relations.
\newblock In \emph{Proceedings of the 2013 Conference on Empirical Methods in
  Natural Language Processing}, pages 681--692, Association for Computational
  Linguistics.

\bibitem[{Lewis and Steedman(2014)}]{Lewis14a*ccg}
Lewis, Mike and Mark Steedman. 2014.
\newblock A* {CCG} parsing with a supertag-factored model.
\newblock In \emph{In Proceedings of the Conference on Empirical Methods in
  Natural Language Processing}.

\bibitem[{Moshier and Rounds(1987)}]{moshier1987succinctness}
Moshier, M.~Drew and William~C. Rounds. 1987.
\newblock On the succinctness properties of unordered context-free grammars.
\newblock In \emph{Proceedings of the 25th Annual Meeting of the Association
  for Computational Linguistics (ACL)}, pages 112--116, Stanford, CA, USA.

\bibitem[{Papadimitriou(1994)}]{papadimitriou1994computational}
Papadimitriou, Christos~H. 1994.
\newblock \emph{Computational Complexity}.
\newblock Addison-Wesley.

\bibitem[{Rimell, Clark, and Steedman(2009)}]{rimell2009unbounded}
Rimell, Laura, Stephen Clark, and Mark Steedman. 2009.
\newblock Unbounded dependency recovery for parser evaluation.
\newblock In \emph{Proceedings of the 2009 Conference on Empirical Methods in
  Natural Language Processing (EMNLP)}, pages 813--821, Singapore.

\bibitem[{Ristad(1986)}]{ristad1986computational}
Ristad, Eric~S. 1986.
\newblock The computational complexity of current {GPSG} theory.
\newblock In \emph{Proceedings of the 24th Annual Meeting of the Association
  for Computational Linguistics (ACL)}, pages 30--39, New York, USA.

\bibitem[{Satta(1992)}]{satta1992recognition}
Satta, Giorgio. 1992.
\newblock Recognition of linear context-free rewriting systems.
\newblock In \emph{Proceedings of the 30th Annual Meeting of the Association
  for Computational Linguistics (ACL)}, pages 89--95, Newark, DE, USA.

\bibitem[{Schabes(1990)}]{schabes1990mathematical}
Schabes, Yves. 1990.
\newblock \emph{Mathematical and Computational Aspects of Lexicalized
  Grammars}.
\newblock Ph.D. thesis, University of Pennsylvania, Philadelphia, USA.

\bibitem[{Seki et~al.(1991)Seki, Matsumura, Fujii, and
  Kasami}]{seki1991multiple}
Seki, Hiroyuki, Takashi Matsumura, Mamoru Fujii, and Tadao Kasami. 1991.
\newblock On {Multiple Context-Free Grammars}.
\newblock \emph{Theoretical Computer Science}, 88(2):191--229.

\bibitem[{Steedman(2000)}]{steedman2000syntactic}
Steedman, Mark. 2000.
\newblock \emph{The Syntactic Process}.
\newblock MIT Press.

\bibitem[{Steedman and Baldridge(2011)}]{steedman2011combinatory}
Steedman, Mark and Jason Baldridge. 2011.
\newblock {Combinatory Categorial Grammar}.
\newblock In Robert~D. Borsley and Kersti B{\"o}rjars, editors,
  \emph{Non-Transformational Syntax: Formal and Explicit Models of Grammar}.
  Blackwell, chapter~5, pages 181--224.

\bibitem[{Vijay-Shanker and Joshi(1985)}]{vijayshanker1985some}
Vijay-Shanker, K. and Aravind~K. Joshi. 1985.
\newblock Some computational properties of {Tree Adjoining Grammars}.
\newblock In \emph{Proceedings of the 23rd Annual Meeting of the Association
  for Computational Linguistics (ACL)}, pages 82--93, Chicago, USA.

\bibitem[{Vijay-Shanker and Weir(1990)}]{vijayshanker1990polynomial}
Vijay-Shanker, K. and David~J. Weir. 1990.
\newblock Polynomial time parsing of combinatory categorial grammars.
\newblock In \emph{Proceedings of the 28th Annual Meeting of the Association
  for Computational Linguistics (ACL)}, pages 1--8, Pittsburgh, USA.

\bibitem[{Vijay-Shanker and Weir(1993)}]{vijayshanker1993parsing}
Vijay-Shanker, K. and David~J. Weir. 1993.
\newblock Parsing some constrained grammar formalisms.
\newblock \emph{Computational Linguistics}, 19(4):591--636.

\bibitem[{Vijay-Shanker and Weir(1994)}]{vijayshanker1994equivalence}
Vijay-Shanker, K. and David~J. Weir. 1994.
\newblock The equivalence of four extensions of context-free grammars.
\newblock \emph{Mathematical Systems Theory}, 27(6):511--546.

\bibitem[{Vijay-Shanker, Weir, and
  Joshi(1987)}]{vijayshanker1987characterizing}
Vijay-Shanker, K., David~J. Weir, and Aravind~K. Joshi. 1987.
\newblock Characterizing structural descriptions produced by various
  grammatical formalisms.
\newblock In \emph{Proceedings of the 25th Annual Meeting of the Association
  for Computational Linguistics (ACL)}, pages 104--111, Stanford, CA, USA.

\bibitem[{Weir and Joshi(1988)}]{weir1988combinatory}
Weir, David~J. and Aravind~K. Joshi. 1988.
\newblock Combinatory categorial grammars: Generative power and relationship to
  linear context-free rewriting systems.
\newblock In \emph{Proceedings of the 26th Annual Meeting of the Association
  for Computational Linguistics (ACL)}, pages 278--285, Buffalo, USA.

\bibitem[{White, Clark, and Moore(2010)}]{white2010generating}
White, Michael, Robert A.~J. Clark, and Johanna~D. Moore. 2010.
\newblock Generating tailored, comparative descriptions with contextually
  appropriate intonation.
\newblock \emph{Computational Linguistics}, 36(2):159--201.

\bibitem[{Zhang and Clark(2011)}]{zhang2011shift}
Zhang, Yue and Stephen Clark. 2011.
\newblock Shift-reduce {CCG} parsing.
\newblock In \emph{Proceedings of the 49th Annual Meeting of the Association
  for Computational Linguistics (ACL)}, pages 683--692, Portland, OR, USA.

\bibitem[{Zhang and Clark(2015)}]{zhang2015discriminative}
Zhang, Yue and Stephen Clark. 2015.
\newblock Discriminative syntax-based word ordering for text generation.
\newblock \emph{Computational Linguistics}, 41(3):503--538.

\end{thebibliography}

\end{document}